%% file: main.tex
\documentclass[acmsmall]{acmart}

\AtBeginDocument{%
  \providecommand\BibTeX{{%
    \normalfont B\kern-0.5em{\scshape i\kern-0.25em b}\kern-0.8em\TeX}}}

\setcopyright{acmlicensed}
\copyrightyear{2024}
\acmYear{2024}
\acmDOI{XXXXXXX.XXXXXXX}

\usepackage{soul}
\input{macros}

\input{util/packages}

\newcommand{\bluetext}[1]{\textcolor{black}{#1}}

\begin{document}

\title{Federated Learning Client Pruning for Noisy Labels}


\author{Mahdi Morafah}
\authornote{Corresponding author (email: mmorafah@ucsd.edu). \\ Accepted to ACM Transactions on Modeling and Performance Evaluation of Computing Systems.}
\email{mmorafah@ucsd.edu}
\affiliation{
  \institution{University of California San Diego}
  \country{USA}
}

\author{Hojin Chang}
\email{hmchang@ucsd.edu}
\affiliation{
  \institution{University of California San Diego}
  \country{USA}
}

\author{Chen Chen}
\email{chen.chen@crcv.ucf.edu}
\affiliation{
  \institution{University of Central Florida}
  \country{USA}
}

\author{Bill Lin}
\email{billlin@ucsd.edu}
\affiliation{
  \institution{University of California San Diego}
  \country{USA}
}

\input{secs/abstract}




\maketitle

\input{secs/introduction}

\input{secs/related_works}

\input{secs/problem}
\input{secs/method}
\input{secs/experiments}

\input{secs/hyperparameters}

\input{secs/conclusion}


\bibliographystyle{ACM-Reference-Format}
\bibliography{refs}

\end{document}

%% file: macros.tex
\usepackage[mathscr]{euscript}
\usepackage{accents}

\DeclareSymbolFont{rsfs}{U}{rsfs}{m}{n}
\DeclareSymbolFontAlphabet{\mathscrsfs}{rsfs}

\newcommand{\E}{\mathbb{E}}

\def\bx{{\boldsymbol x}}

\def\btheta{{\boldsymbol \theta}}

\usepackage{color}

\newcommand{\argmin}[1]{\underset{#1}{\mathrm{argmin}} \ }



%% file: util/packages.tex
\usepackage{inputenc}
\usepackage{geometry}
\usepackage{url}
\usepackage{amsfonts} 
\usepackage{bm}
\usepackage{nicefrac}
\usepackage{microtype}
\usepackage{color}
\usepackage{colortbl}
\usepackage{amssymb}
\usepackage{amsthm}
\usepackage{pifont}
\usepackage{natbib} \setcitestyle{square,sort,comma,numbers}
\usepackage{mathtools}
\usepackage{dirtytalk}
\usepackage{subcaption}
\usepackage{xspace}
\usepackage{forloop}
\usepackage{makecell}
\usepackage{multirow}
\usepackage{algorithm}
\usepackage{algorithmic}
\usepackage[shortlabels]{enumitem}
\usepackage{float}
\usepackage{xcolor}
\usepackage{hyperref}

\graphicspath{ {./images/} }

\definecolor{Gray}{gray}{0.9}
\definecolor{White}{gray}{1.0}
\definecolor{DarkRed} {RGB}{219, 112, 112}
\definecolor{DarkBlue}{RGB}{112, 112, 219}
\definecolor{DarkGray}{RGB}{112, 112, 112}
\definecolor{PaleRed} {RGB}{255, 236, 236}
\definecolor{PaleBlue}{RGB}{236, 236, 255}
\definecolor{PaleGray}{RGB}{236, 236, 236}

%% file: secs/abstract.tex
\begin{abstract}
Federated Learning (FL) enables collaborative model training across decentralized edge devices while preserving data privacy. However, existing FL methods often assume clean annotated datasets, impractical for resource-constrained edge devices. In reality, noisy labels are prevalent, posing significant challenges to FL performance. Prior approaches attempt label correction and robust training techniques but exhibit limited efficacy, particularly under \bluetext{high noise levels}. This paper introduces ClipFL (\textbf{F}ederated \textbf{L}earning \textbf{Cli}ent \textbf{P}runing), a novel framework addressing noisy labels from a fresh perspective. ClipFL identifies and excludes noisy clients based on their performance on a clean validation dataset, tracked using a Noise Candidacy Score (NCS). The framework comprises three phases: pre-client pruning to identify potential noisy clients and calculate their NCS, client pruning to exclude a percentage of clients with the highest NCS, and post-client pruning for fine-tuning the global model with standard FL on clean clients. Empirical evaluation demonstrates ClipFL's efficacy across diverse datasets and noise levels, achieving accurate noisy client identification, superior performance, faster convergence, and reduced communication costs compared to state-of-the-art FL methods. Our code is available at~\url{https://github.com/MMorafah/ClipFL}.
\end{abstract}

%% file: secs/introduction.tex
\section{Introduction}

Edge devices such as IoTs and mobile devices are increasingly ubiquitous, constituting a new computational platform for machine learning. Despite holding vast real-world data, these billions of devices often withhold their data due to privacy concerns. Federated Learning (FL) emerges as a decentralized machine learning paradigm, enabling model training with the collaboration of multiple clients while preserving privacy \cite{mcmahan2017communication, kairouz2021advances}. FL shows promise in enhancing model performance without necessitating data sharing and finds applications across diverse domains \cite{kairouz2021advances, sui2020feded, muhammad2020fedfast, dayan2021federated, hard2018federated}. However, FL encounters significant performance degradation in the presence of data heterogeneity \cite{haddadpour2019convergence, li2019feddane, zhao2018federated}. Recent advancements introduce FL optimizers tailored to address data heterogeneity, achieving \bluetext{faster} convergence \cite{li2020federated, karimireddy2020scaffold, wang2020tackling, reddi2020adaptive}.

Despite these advancements, the majority of prior works operate under the assumption of accurately labeled and clean client data. In practice, however, acquiring precisely annotated clean datasets is arduous and resource-intensive, especially for edge devices lacking ample resources. Consequently, the labels in their datasets often contain noise. Unfortunately, FL experiences substantial performance degradation when confronted with noisy clients \cite{yang2022robust}. Thus, developing FL frameworks resilient to noisy labels is imperative. Several prior works propose methods to identify and rectify noisy samples using the global model's predictions \cite{xu2022fedcorr, yang2022robust}. For instance, FedCorr \cite{xu2022fedcorr} presents a multi-stage FL framework that identifies noisy clients by measuring the local intrinsic dimensionality (LID) and corrects noisy labels using global model predictions. \bluetext{In order to control the negative impact of noisy client in achieving a well-trained reliable global model prior to label correction stage, they incorporate several techniques including client fraction scheduling scheme and local proximal regularization with mix-up.} However, this approach heavily relies on a well-performing global model, which is challenging to obtain in the presence of data heterogeneity and noisy clients, leading to inaccurate label correction and suboptimal performance. Other approaches aim to mitigate the impact of noisy clients through client weighting strategies and robust local training methods \cite{fang2022robust, yiqiang2020focus, jiang2022towards}. For example, RHFL \cite{fang2022robust} utilizes symmetric cross-entropy loss during local training and introduces a client confidence re-weighting scheme to counteract the adverse effects of noisy labels during collaborative learning. However, these methods demonstrate limited efficacy and poor convergence, especially under high noise levels.

In this paper, we address the challenge of noisy labels in FL by adopting an alternative approach compared to prior works. Rather than mitigating or correcting noisy clients, we propose a novel \emph{``\textbf{F}ederated \textbf{L}earning \textbf{Cli}ent \textbf{P}runing''} framework called \textbf{ClipFL}. ClipFL identifies noisy clients and excludes them from the FL training process. Our framework comprises of three phases. In the initial phase, we identify noisy clients based on their performance on a clean validation dataset, tracking the frequency of noisy identifications as \emph{Noise Candidacy Score (NCS)} of each client. To mitigate the negative impact of noisy clients during server-side model aggregation, we only aggregate the top-$m$ clients with the highest validation accuracy. Subsequently, in the second phase, we prune $p\%$ of the clients with the highest noise candidacy score (NCS) in a one-shot manner. Finally, in the third stage, we conduct standard FL on the remaining clean clients to further refine the global model.

\begin{figure}[t]
    \centering
    \begin{minipage}{0.53\textwidth}
         \vspace{6pt}
         \includegraphics[width=1.0\textwidth]{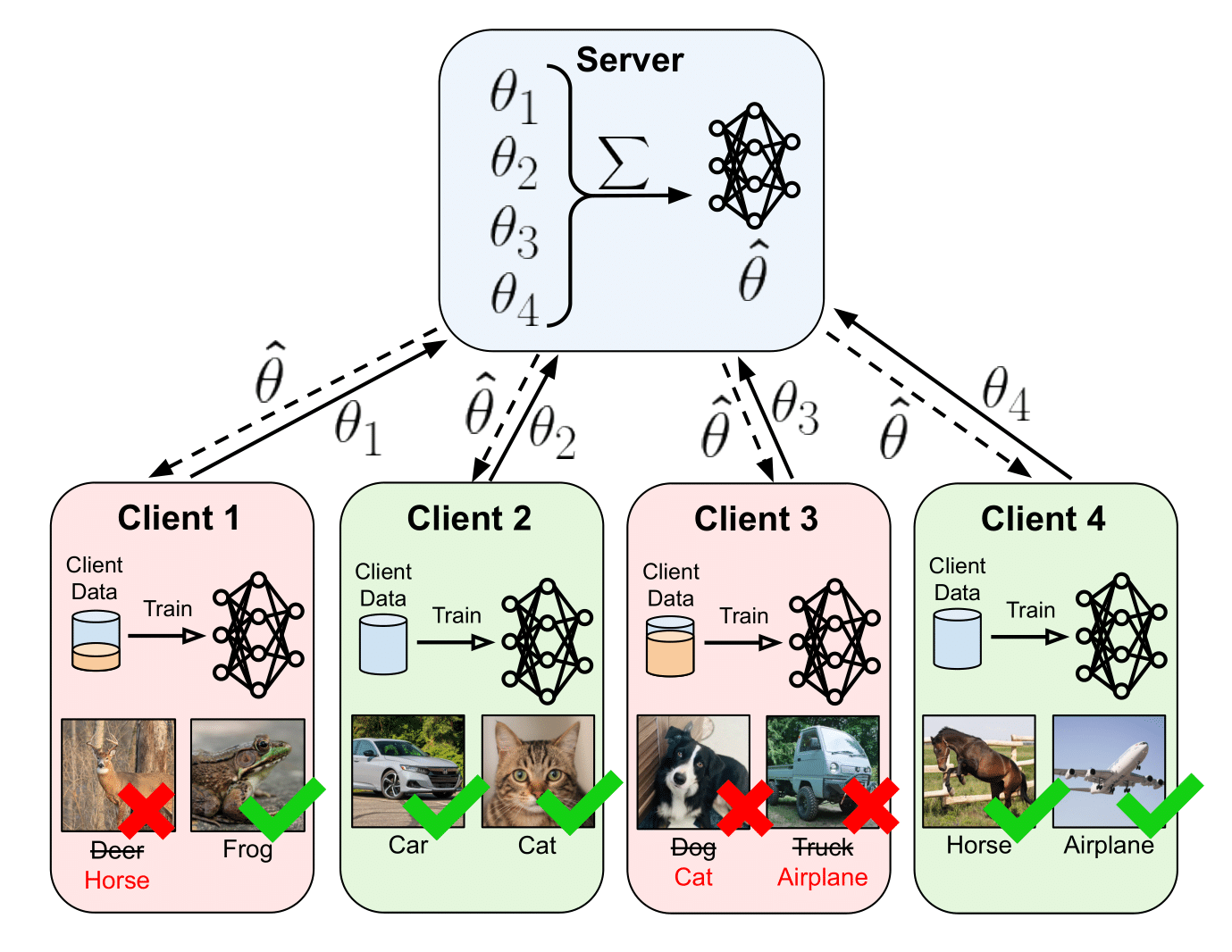}
         \subcaption{Prior works typically involve incorporating noisy clients into FL and addressing their negative impact through methods such as correcting noisy labels, re-weighting noisy clients, or designing robust local training methods.}
    \end{minipage}%
    \hspace{0.01\textwidth}
    \begin{minipage}{0.4\textwidth}
        \centering
        \vspace{5pt}
         \includegraphics[width=0.76\textwidth]{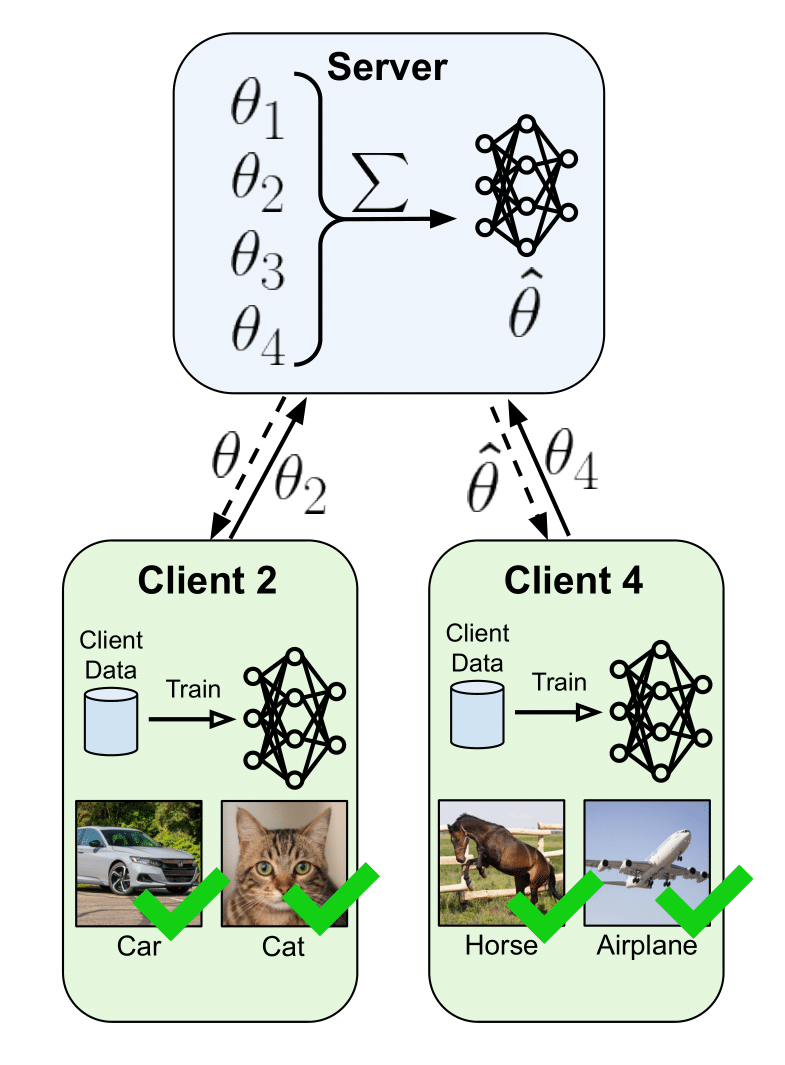}
         \subcaption{In contrast to prior works, ClipFL introduces the Noise Candidacy Score (NCS) for each client, enabling robust identification of noisy clients from the FL process.}
    \end{minipage}%
    \caption{Comparison between our approach and prior works: Prior works often face challenges related to poor convergence and reliance on a well-performing global model as a pseudo-labeler that corrects noisy labels, which can be difficult to obtain, especially in the presence of high noise levels and data heterogeneity (see Section~\ref{RQ2}). In contrast, ClipFL offers consistent performance improvements and reduces communication costs by robustly identifying noisy clients and excluding them from the FL process.}
    \label{fig:advantage}
\end{figure}

We validate ClipFL on diverse datasets with varying noise levels for both IID and Non-IID data partitions, yielding several key observations: (1) ClipFL accurately identifies noisy clients with at least 80\% accuracy on most of the cases. (2) ClipFL outperforms state-of-the-art (SOTA) FL optimizers without excluding noisy clients. (3) ClipFL surpasses existing FL methods addressing noisy labels, underscoring the effectiveness of our excluding approach. (4) ClipFL demonstrates faster convergence and reduced communication costs compared to both SOTA vanilla FL optimizers and FL methods designed to counteract noisy labels. We make our code publicly available at~\url{https://github.com/MMorafah/ClipFL}.

\textbf{Contribution.} Our contributions are threefold:
\begin{itemize}
\item We introduce ClipFL, a novel federated learning client excluding method to address the challenge of noisy labels in FL.
\item ClipFL distinguishes clean clients from noisy ones through evaluation on a clean validation dataset and aggregates only the top-performing clients to mitigate the impact of noisy clients.
\item We empirically evaluate ClipFL across different datasets with varying noise levels for both IID and Non-IID data partitions, demonstrating significant performance improvements and reduced communication costs over SOTA FL methods.
\end{itemize}

\textbf{Organization.} The rest of the paper is organized as follows: In Section~\ref{sec:related}, we discuss related works. Section~\ref{sec:background} presents the background and problem formulation. Our proposed method is introduced in Section~\ref{sec:methodology}. Section~\ref{sec:experiments} details our experimental results. We discuss our implementation and hyperparameters in Section~\ref{sec:imp_hp}. Finally, we conclude our work in Section~\ref{sec:conclusion}.

%% file: secs/related_works.tex
\section{Related Works} \label{sec:related}
\noindent\textbf{FL with Non-IID data.} Federated learning encounters significant performance degradation and convergence challenges in the presence of data heterogeneity~\cite{tan2022towards, zhao2018federated}. To address this issue, numerous strategies have been explored. Some studies propose adjustments to local training methodologies to improve performance~\cite{li2020federated, li2021model, mendieta2022local}. For instance, FedProx~\cite{li2020federated} introduces a proximal term during local training to penalize parameter differences from the global model. MOON~\cite{li2021model} employs contrastive learning to preserve global knowledge during local training.

Other works focus on alternative FL optimization procedures with improved convergence guarantees~\cite{wang2020tackling, karimireddy2020scaffold, karimireddy2020mime, reddi2020adaptive, haddadpour2019convergence}. For example, FedNova~\cite{wang2020tackling} proposes a normalized weighted aggregation technique to address the objective inconsistency caused by data heterogeneity. FedOpt~\cite{reddi2020adaptive} introduces federated versions of adaptive optimizers such as Adam, AdaGrad, and Yogi, demonstrating significant performance improvements. Scaffold~\cite{karimireddy2020scaffold} focuses on correcting local updates and mitigating client drift by introducing control variates. Additionally, ensemble distillation techniques~\cite{lin2020ensemble, sattler2021fedaux} and neuron matching approaches~\cite{wang2020federated, singh2020model, li2022federated, liu2022deep} have been proposed to address challenges associated with non-IID data in federated learning.
\\

\noindent\textbf{FL with noisy labels.}
Several approaches have been explored in prior works to tackle the challenge of noisy labels in federated learning\bluetext{~\cite{liang2023fednoisy}}. 
Some methods focus on identifying noisy clients and correcting noisy samples via the global model's predictions~\cite{xu2022fedcorr, yang2022robust}. For instance, FedCorr~\cite{xu2022fedcorr} introduces a multi-stage FL framework that identifies noisy clients by assessing the local intrinsic dimensionality (LID) of local model prediction subspaces. It identifies noisy labels by measuring local cross-entropy loss and correcting them with global model predictions. \bluetext{To ensure a reliable and well-performing global model, FedCorr employs an adaptive local proximal term added to the local training loss, mix-up data augmentation, and a client fraction scheduling scheme, where a smaller number of clients are selected without replacement. } ~\citet{yang2022robust} form global class-wise feature centroids based on samples with relatively small losses to address noisy labels and correct less confident samples using the global model. However, these approaches heavily depend on a well-performing global model as a pseudo-labeler reference, which is challenging to obtain in the presence of data heterogeneity and noisy clients, leading to inaccurate label correction. \bluetext{Despite typically introducing noise reduction procedures prior to label correction, these methods still face significant challenges in achieving a well-trained, reliable global model. \citet{zeng2022clc} further highlight this issue and propose CLC (Consensus-based Label Correction), which aims to correct noisy labels using a consensus method among FL participants. CLC aggregates class-wise information from all participants to estimate the likelihood of label noise in a privacy-preserving manner, using consensus-defined class-wise thresholds and a margin tool to correct noisy labels.}

Other strategies aim to mitigate the adverse effects of noisy clients by proposing client weighting strategies and employing local training methods robust to noisy labels~\cite{fang2022robust, yiqiang2020focus, jiang2022towards}. For example, RHFL~\cite{fang2022robust} utilizes symmetric cross-entropy loss during local training and introduces a client confidence re-weighting scheme to alleviate the adverse effects of noisy labels during collaborative learning. In~\cite{yiqiang2020focus}, a credibility-weighted aggregation scheme is proposed, where weights are computed based on the cross-entropy between the predictions of a server-side model trained on a benchmark dataset and the predictions made by local models on their respective local data. However, this approach necessitates the existence of a well-performing benchmark model.~\citet{jiang2022towards} introduce a local self-regularization technique which comprises of a Mixup prediction loss that prevents over-fitting on noisy labels and a self-distillation loss that explicitly regularizes the model output discrepancy between original and augmented input images. Additional approaches such as client selection~\cite{yang2021client} and relevant data subset selection~\cite{tuor2021overcoming} have also been proposed to mitigate the impact of noisy labels in federated learning.

However, despite these efforts, prior works often suffer from poor convergence and limited performance, particularly when dealing with large noise levels (see Section~\ref{RQ2}). In contrast, we tackle this problem from a different perspective by identifying noisy clients and pruning them from the federation. \bluetext{It is important to note that, in this context, the removal of noisy data does not necessarily result in the loss of useful information but rather in the elimination of detrimental noisy data that would otherwise degrade performance.} Our approach offers a novel solution \bluetext{and new insights} to the problem of noisy labels in federated learning, overcoming the limitations of previous methods, and reducing the communication cost.

%% file: secs/problem.tex
\section{Background and Problem Formulation} \label{sec:background}
\subsection{Background}
\textbf{Vanilla Federated Averaging (FedAvg).} Let's consider a scenario where we have a collection of clients $S$, each equipped with a local dataset $D_k$. These clients collaborate in a federated learning framework to train a machine learning model $f(\cdot; \btheta)$ over the whole dataset $D = \bigcup_{k=1}^{|S|} D_k$, all while maintaining the privacy of their local data. The vanilla FedAvg procedure is as follows; at each communication round $r$, a fraction $C \in (0, 1]$ of clients is randomly selected, denoted as $S_r$. These selected clients receive the current global model parameters $\btheta^r$ and conduct local training, subsequently sending back their updated parameters $\{{\btheta}_k\}_{k \in S_r}$. The server then aggregates these locally updated parameters to update the global model, defined as $\btheta^{r+1} = \sum_{k \in S_r} \frac{|D_k|}{\sum_{k \in S_r} |D_k|} {\btheta}_k$. This iterative process continues for a total of $R$ communication rounds. \\

\begin{figure}[t]
    \centering
    \begin{minipage}{0.4\textwidth}
    \centering
        $\begin{bmatrix}
1-\mu & \frac{\mu}{K-1} & \cdots & \frac{\mu}{K-1} \\
\frac{\mu}{K-1} & 1-\mu &  & \frac{\mu}{K-1} \\
\vdots &  & \ddots & \vdots \\
\frac{\mu}{K-1} & \frac{\mu}{K-1} & \cdots & 1-\mu \\
\end{bmatrix}$\\
\vspace{0.55in}
        \subcaption{Symmetric noise transition matrix}
    \end{minipage}%
    \hspace{0.01\textwidth}
    \begin{minipage}{0.25\textwidth}
    \centering
        \includegraphics[width=\textwidth]{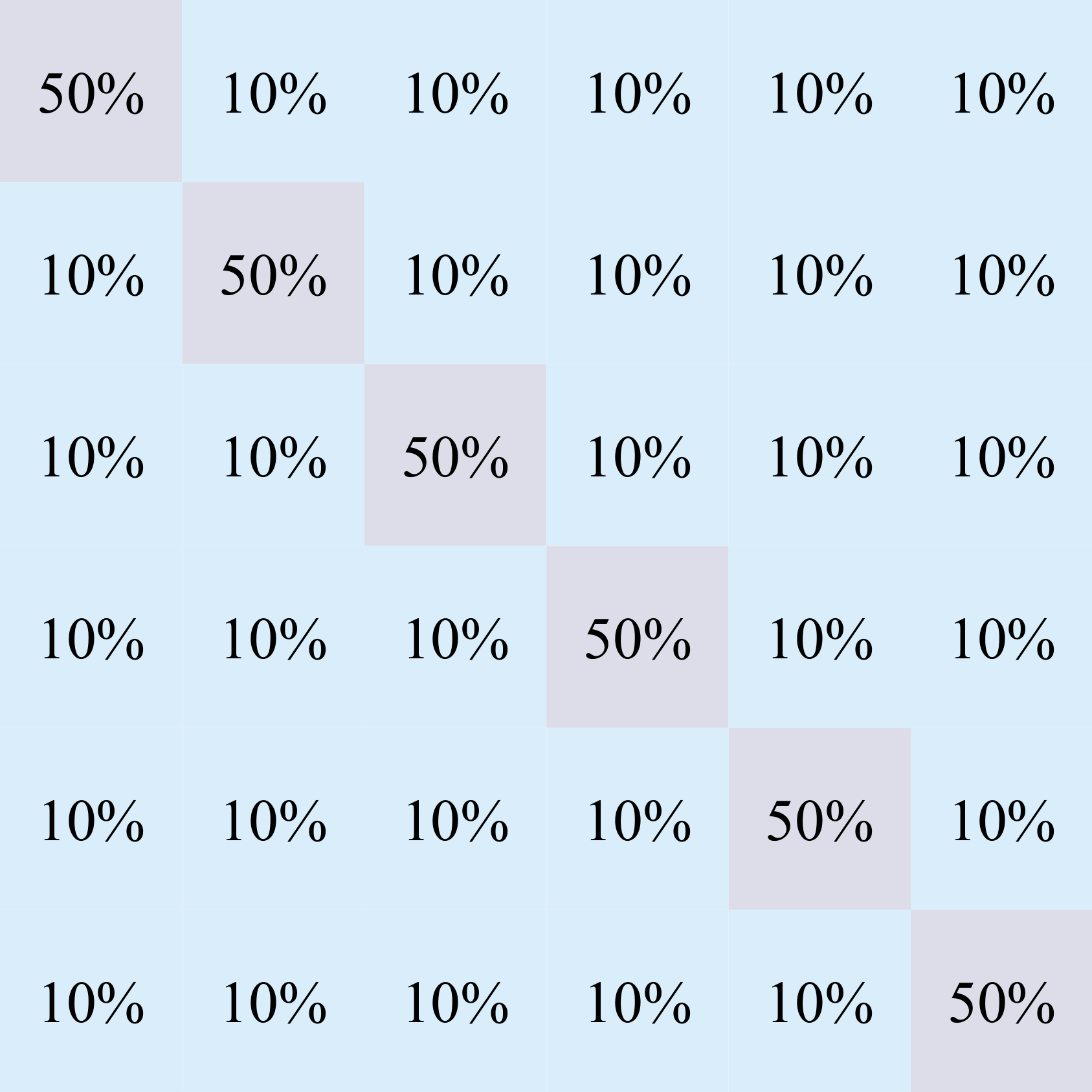}\\
        \subcaption{$\mu=0.5$}
    \end{minipage}%
    \hspace{0.01\textwidth}
    \begin{minipage}{0.25\textwidth}
    \centering
        \includegraphics[width=\textwidth]{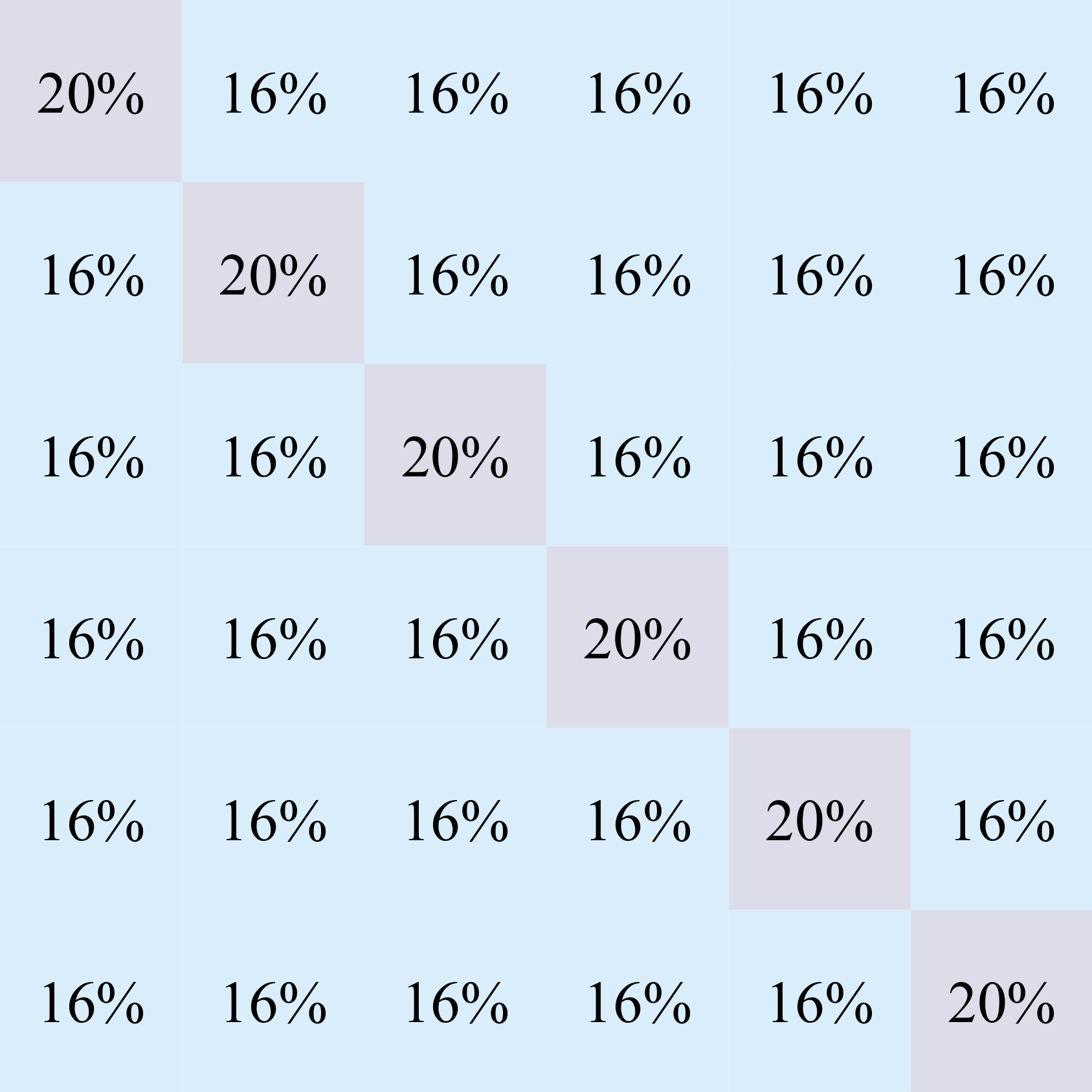}\\
        \subcaption{$\mu=0.8$}
    \end{minipage}%
    \caption{Symmetric Noise Transition Matrix $T$. (a) General structure of matrix $T$. (b) Example of matrix $T$ for 6 classes ($K=6$) with noise level $\mu=0.5$. (c) Example of matrix $T$ for 6 classes ($K=6$) with noise level $\mu=0.8$.}
    \label{fig:noise_matrix}
\end{figure}

\noindent\textbf{Label Noise.} Let's consider a dataset $D: \mathcal{X} \times \mathcal{Y}$, where $\mathcal{X}$ represents the feature space and $\mathcal{Y}$ denotes the set of actual labels, comprising $K$ distinct labels. To introduce manual corruption into the dataset, we utilize a noise transition matrix following~\cite{wei2020combating}. This matrix encapsulates the probabilities of true class labels being corrupted or mislabeled as other class labels during the labeling process. Formally, let $\mathbf{T}$ denote the symmetric transition matrix, where $T_{ij}$ represents the probability of an instance with true class $i$ being observed as class $j$. The matrix $\mathbf{T}$ satisfies the conditions $T_{ij} = T_{ji}$ and $\sum_{j} T_{ij} = 1, \; \forall i$. Two common approaches for label noise modeling are based on utilizing either a symmetric or asymmetric noise transition matrix~\cite{han2020survey, song2022learning}. In the case of symmetric label noise modeling, to construct a noise transition matrix $\mathbf{T}$ with a noise level of $\mu$, we assign the diagonal elements to $1-\mu$, while the off-diagonal elements are set to $\frac{\mu}{K-1}$ to maintain symmetry. Figure~\ref{fig:noise_matrix} illustrates an example of a noise transition matrix $\mathbf{T}$ for symmetric label noise modeling. \bluetext{In the asymmetric noise matrix, the probability of a label being corrupted to another label is not uniform; it depends on both the source and target classes. Therefore, asymmetric noise matrix can have a variety of structures. In the literature symmetric noise modeling has been widely adopted due to its challenging nature and  a more standardize and universal benchmark for evaluation.}

\subsection{Problem Formulation}
Suppose a federated learning setting with $N$ clients where each client $k$'s local private data is denoted by $D_k$. Let $S$ denote the set of entire clients. Moreover, the set of noisy clients is $S_n$, and the complement of set $S_n$ is the set of clean clients: $S_c = S\setminus S_n$ and $S = S_c \cup S_n$. The objective is solving the following minimization problem:
\begin{align} \label{eq:gfl}
    \widehat \btheta = \argmin {\btheta} \sum_{i=1}^N \E_{(\bx,y)\sim D_i}[\ell(f(\bx; \btheta), y)],
\end{align}
where $\ell$ denotes the appropriate loss function (e.g. cross-entropy), and $f$ denotes the neural network model.

%% file: secs/method.tex
\section{Proposed Method} \label{sec:methodology}

\begin{figure}[t]
    \centering
    \includegraphics[width=1.0\textwidth]{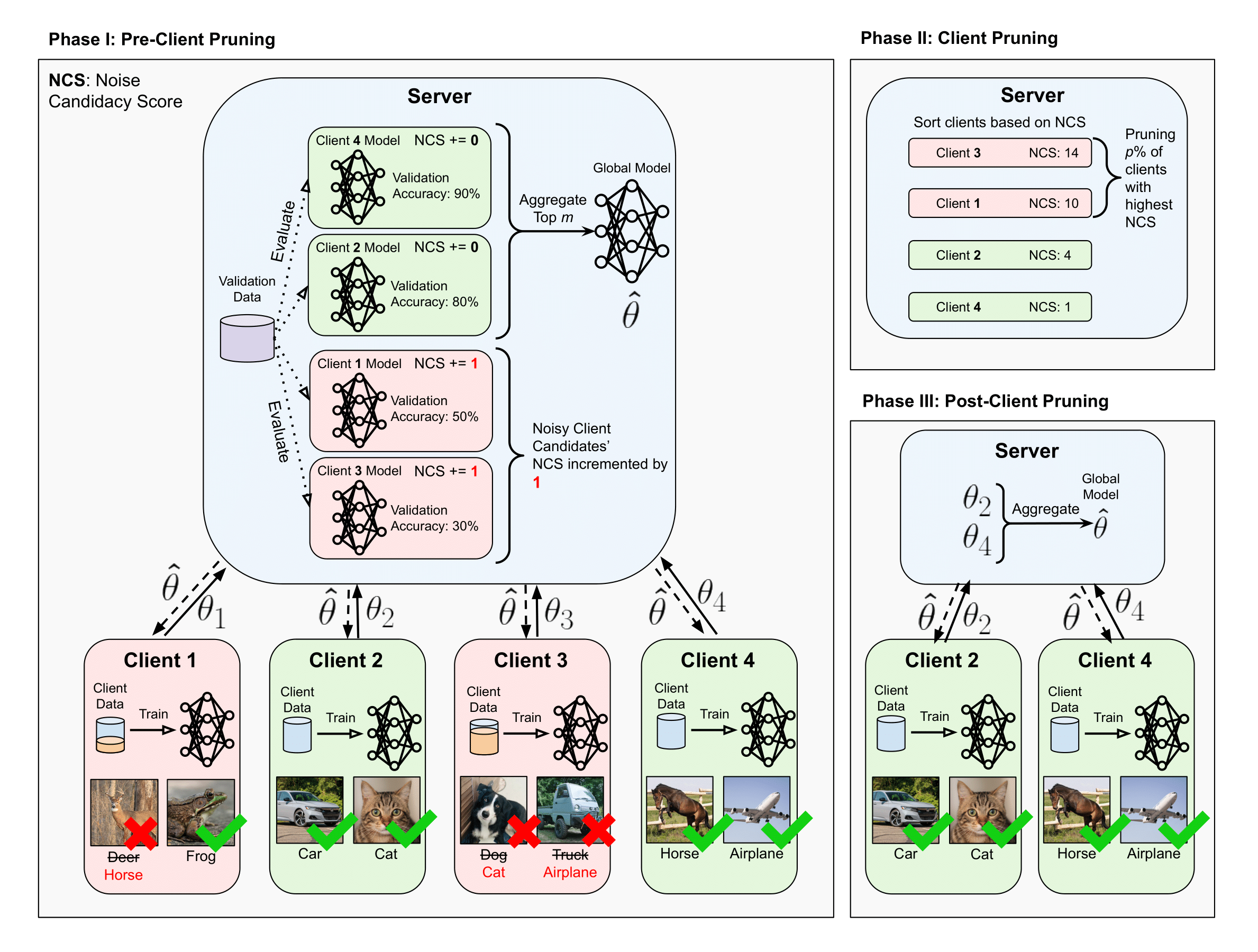}
    \caption{Overview of ClipFL: A federated learning framework addressing the challenge of a subset of clients withholding mislabeled samples. ClipFL operates through three phases: Phase I involves evaluating client models at the end of each round using a clean validation dataset. The top-$m$ highest performing clients are identified as potential clean clients, while the rest are flagged as potential noisy clients. To mitigate negative impacts from noisy clients, ClipFL updates the global model using only clean client candidates. We also introduce the \emph{Noise Candidacy Score (NCS)} for each client, reflecting the number of times they are identified as noisy throughout the FL process in this phase. In Phase II, we prune $p\%$ of the clients with the highest Noise Candidacy Scores from participating in FL. Finally, in Phase III, we refine the global model by performing standard FL on the remaining clean clients.}
    \label{fig:method}
\end{figure}


\subsection{Overview}
In this section, we introduce ClipFL, a method designed to address the challenges of federated learning with noisy labels. ClipFL operates within the standard federated learning framework (as described in Section~\ref{sec:background}) and does not necessitate any warm-up federated learning training rounds. Illustrated in Figure~\ref{fig:method}, ClipFL comprises of three distinct phases:
\begin{itemize}
    \item \textbf{Phase I (Pre-Client Pruning):}  At the end of each round, the server evaluates the model of each client locally trained that round using a clean validation dataset to identify potential noisy client candidates. Specifically, we designate the top-$m$ highest-performing clients as clean, while the remaining clients are considered potentially noisy client candidates. To mitigate the potential negative impact of these identified noisy client candidates during server-side aggregation, only models from the identified potentially clean clients are aggregated. Furthermore, we record the number of times each client is identified as a noisy client candidate over multiple rounds during the pre-client pruning phase, defined as the Noise Candidacy Score of each client. 
    \item \textbf{Phase II (Client Pruning):} In the second phase, we utilize the noise candidacy scores of clients as a measure of identifying noisy clients. We sort the clients based on their noise candidacy scores in descending order and prune $p\%$ of the clients with the highest noise candidacy scores from the entire federation, as captured during the first phase, in a one-shot manner.
    \item \textbf{Phase III (Post-Client Pruning):} In the third phase, we refine the global model further by conducting normal federated learning on the remaining clean clients after pruning.
\end{itemize} 

In the following sections, we delve into the details of each step of the ClipFL method.

\begin{algorithm}
\caption{ClipFL} \label{alg:clipfl}
\begin{algorithmic}[1]
\REQUIRE initial model ($\bm{\theta}^0$), pre-pruning rounds ($T_{pre}$), post-pruning rounds ($T_{post}$), total number of clients ($N$), sampling rate ($C$), client pruning percentage ($p$), number of clients to be considered as clean in pre-pruning phase ($m$), global validation set ($D^{val}$), local client training algorithm ($\mathtt{ClientUpdate}$\footnotemark[1]), federated model aggregation algorithm ($\mathtt{ModelFusion}$\footnotemark[2]) \\
\hspace{-0.32cm}\textbf{Server Executes:} \\
\STATE Initialize the server model with $\bm{\theta}^0$ \\
\STATE $NCS \leftarrow$ noise candidacy score (NCS) of each client initialized to 0 \\
\colorbox{PaleRed}{
\hspace{-0.2cm}\begin{minipage}{0.94\textwidth}
\textcolor{DarkRed}{$\bm{\mathtt{Phase \:I: Pre- Client \: Pruning}}$}
\FOR {each round $t = 0, 1, \ldots, T_{pre}-1$}
    \STATE $S^t \leftarrow$ Randomly select $\lfloor \vert S \vert \cdot C \rfloor$ clients from the set of available clients $S$
    \FOR {each client $k$ in $S^t$ \textbf{in parallel}}
        \STATE $\btheta_k \leftarrow \mathtt{ClientUpdate}(k, \btheta^t)$
    \ENDFOR
    \STATE $A \leftarrow$ evaluate each client model $\btheta_k$ on validation dataset $D^{val}$ and capture the accuracy
    \STATE $S^t \leftarrow \mathtt{sort}(S^t, A)$ \COMMENT{\textit{sort clients $S^t$ based on validation accuracy $A$ in descending order}}
    \STATE $\hat S^t_c = \{ S^t[i] \mid 1 \leq i \leq m \}$ \COMMENT{\textit{identifying top-m clients as potential clean clients}}
    \STATE $\hat S^t_n = S^t \setminus \hat S^t_c$ \COMMENT{\textit{identifying the rest of $\vert S^t \vert - m$ clients as potential noisy clients}}
    \STATE $\text{NCS}[i] \; += 1, \: \forall i\in \hat S_n^t$ \COMMENT{\textit{updating noise candidacy score of identified noisy clients}}
    \STATE $\btheta^{t+1} \leftarrow \mathtt{ModelFusion}(\{\btheta_l\}_{l\in S_c^t})$
\ENDFOR
\end{minipage}
}
\colorbox{PaleBlue}{
\hspace{-0.2cm}\begin{minipage}{0.94\textwidth}
\textcolor{DarkBlue}{$\bm{\mathtt{Phase \:II: Client \: Pruning}}$}
\STATE $S \leftarrow \text{sort}(S, \text{NCS})$ \COMMENT{\textit{sort all clients $S$ based on their noise candidacy scores in descending order}}
\STATE $\hat S_n \leftarrow \{ S[i] \mid 1 \leq i \leq \left\lfloor p \cdot \vert S \vert \right\rfloor \}$ \COMMENT{\textit{identifying $p\%$ of clients as noisy}}
\STATE $S \leftarrow S \setminus \hat S_n$ \COMMENT{\textit{updating the set of clients by removing noisy clients}}
\end{minipage}
}
\colorbox{PaleGray}{
\hspace{-0.2cm}\begin{minipage}{0.94\textwidth}
\textcolor{DarkGray}{$\bm{\mathtt{Phase \:III: Post-Client \: Pruning}}$}
\FOR {each round $t = T_{pre}, T_{pre}+1, \ldots, T_{pre}+T_{post}-1$}
    \STATE $S^t \leftarrow$ Randomly select $\lfloor \vert S \vert \cdot C \rfloor$ clients from the set of available clients $S$
    \FOR {each client $k$ in $S^t$ \textbf{in parallel}}
        \STATE $\btheta_k \leftarrow \mathtt{ClientUpdate}(k, \btheta^t)$
    \ENDFOR
    \STATE $\btheta^{t+1} \leftarrow \mathtt{ModelFusion}(\{\btheta_l\}_{l\in S^t})$
\ENDFOR
\end{minipage}
}
\end{algorithmic}
\end{algorithm}

\subsection{Phase I: Pre-Client Pruning}
The primary objective of this phase is to identify noisy clients in the normal federated learning process. At each round $t$, a randomly selected set of clients $S^t$ receives the global model $\btheta^t$. After locally updating their models on their respective local datasets, the clients send back their updated models $\{\btheta_k\}_{k\in S^t}$ to the server. To identify potential noisy clients, the server evaluates each received client $k$'s ($k \in S^t$) model on a clean validation dataset $D^{val}$ and captures the accuracy $A_k$, denoted as $A_k = \text{Eval}(\btheta_k, D^{val})$. Clean clients typically exhibit higher accuracy compared to noisy ones. Therefore, we first sort the clients $S^t$ based on their validation accuracy $A$ as follows:
\begin{align}
S^t = \text{sort}(S^t, A).
\end{align}
Next, we consider $m$ clients with the highest validation accuracy as clean clients ($S^t_c$), as follows:

\noindent\footnotetext[1]{ClipFL is a versatile modular technique that can be added on to different baselines. The specific algorithm details of $\mathtt{ClientUpdate}$ varies by utilizing different FL optimizers. For example, if FedAvg is used with ClipFL, $\mathtt{ClientUpdate}$ would be just normal SGD training.}
\noindent\footnotetext[2]{The model aggregation details vary by adopting different FL optimizers. For example, if FedAvg is used with ClipFL, $\mathtt{ModelFusion}$ would be taking the average of locally trained models.}
\begin{align}
\hat S^t_c = \{ S^t[i] \mid 1 \leq i \leq m \},
\end{align}
and the remaining $\lvert S_t \rvert - m$ clients are considered noisy clients ($S^t_n$), as follows:
\begin{align}
\hat S^t_n = S^t \setminus \hat S^t_c.
\end{align}

To mitigate any negative impact from noisy clients on the server-side model aggregation and to prevent corruption of clean clients, only the identified clean clients are used for aggregation for the global model's update. This process iterates for $T_{pre}$ rounds. 

This phase serves as a crucial pre-processing step, allowing us to identify and isolate noisy clients candidates early in the federated learning process. By prioritizing clean clients candidates for model aggregation, we aim to enhance the quality of the global model and improve overall performance.

\bluetext{\textbf{Remark.} The use of public datasets, specifically validation datasets for hyper-parameter tuning and methodology design, is a common practice in both the centralized machine learning literature and the federated learning literature~\cite{khodak2021federated, li2021hermes, li2021fedmask, vahidian2021personalized, lin2020ensemble}. Furthermore, in practice, there are several organizations in various fields, such as the NIH and WHO in the healthcare domain, that maintain the quality and annotation of publicly available datasets. Additionally, thanks to recent advancements in generative AI, high-quality datasets can be synthesized for different purposes. }

\subsection{Phase II: Client Pruning}
We observe that relying solely on validation accuracy for identifying noisy clients may not be sufficiently accurate, especially in scenarios with fluctuating client performance, such as data heterogeneity or early stages of federated training. For instance, in the early training stages, clean clients may exhibit lower performance due to insufficient training, leading to misidentification of said clean clients as noisy. Similarly, in the presence of data heterogeneity, low performance may stem from data variations rather than actual noise.

Therefore, it is imperative to devise a robust metric for identifying noisy clients that can accommodate clients' performance fluctuations. To address this, we introduce the noise candidacy score ($\text{NCS}$) for each client, which serves as a more robust measure for identifying noisy clients. Specifically, we define the noise candidacy score $\text{NCS}$ of each client $i$ as the number of times the client is identified as a noisy candidate during the $T_{pre}$ federation training rounds in phase I:
\begin{align}
\text{NCS}[i] = \text{Number of times client } i \text{ is identified as a noisy candidate}.
\end{align}
The intuition behind the NCS lies in capturing the historical behavior of clients over the $T_{pre}$ rounds conducted in Phase I, providing a robust basis for pruning decisions. 

We leverage the noise candidacy scores $\text{NCS}$ to robustly identify noisy clients. Initially, we sort all clients ($S$) based on their noise candidacy scores $\text{NCS}$ in descending order:
\begin{align}
S = \text{sort}(S, \text{NCS}).
\end{align}
Next, we select $p\%$ of the clients with the highest noise candidacy scores as the actual noisy clients:
\begin{align}
\hat S_n = \{ S[i] \mid 1 \leq i \leq \left\lfloor p \cdot \vert S \vert \right\rfloor \},
\end{align}
where $\left\lfloor \cdot \right\rfloor$ denotes the floor operation. Subsequently, these identified noisy clients are pruned from the entire federation training process in a one-shot manner at the start of round $T_{pre}$:
\begin{align}
S = S \setminus \hat S_n.
\end{align}
As a result, the total number of clients available in the federation process becomes $N - p \cdot N$.


\begin{figure}
    \centering
    \begin{minipage}{0.48\textwidth}
    \centering
        \begin{minipage}{0.48\textwidth}
        \includegraphics[width=\textwidth]{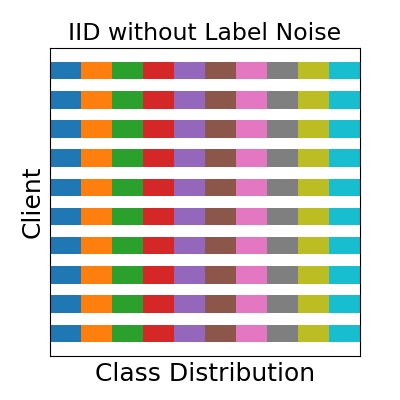}\\
        \end{minipage}
        \begin{minipage}{0.48\textwidth}
        \includegraphics[width=\textwidth]{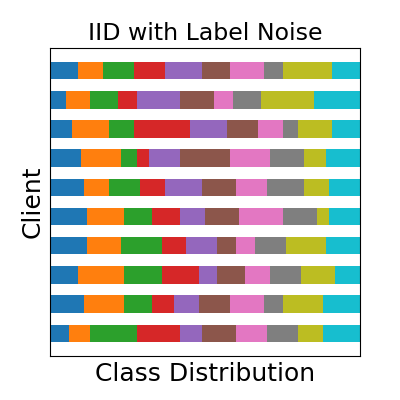}\\
        \end{minipage}
        \subcaption{IID} \label{class_dist_iid}
    \end{minipage}%
    \hspace{0.03\textwidth}
    \begin{minipage}{0.48\textwidth}
    \centering
        \begin{minipage}{0.48\textwidth}
        \includegraphics[width=\textwidth]{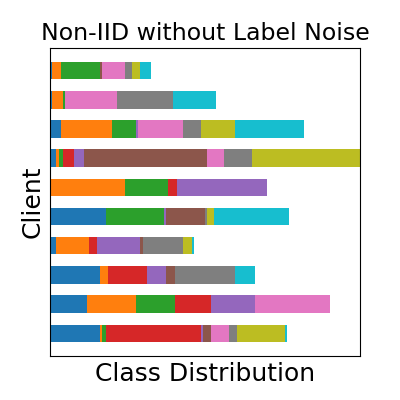}\\
        \end{minipage}
        \begin{minipage}{0.48\textwidth}
        \includegraphics[width=\textwidth]{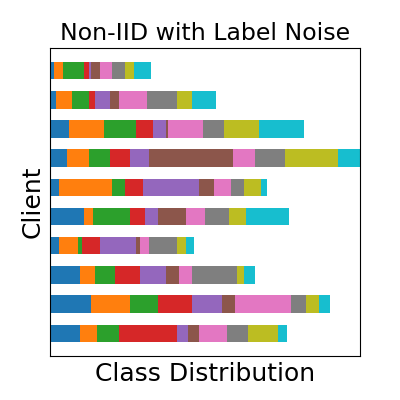}\\
        \end{minipage}
        \subcaption{Non-IID} \label{class_dist_niid}
    \end{minipage}%
    \caption{This figure illustrates an example of the class distribution of clients with and without label noise for both IID and Non-IID data partitioning scenarios. We consider 10 clients and a synthetic dataset with 10 classes (100 samples per class). Each horizontal bar represents the class distribution of one client's dataset, where colors indicate different classes, and bar lengths represent the number of samples. (a) depicts the IID data partitioning, where all samples are distributed IID across clients. (b) represents the Non-IID data partitioning, where samples are distributed according to the Dirichlet distribution with $\alpha=0.5$. We use the symmetric noise transition matrix with $\mu=0.5$ to make each client's data noisy.}
    \label{fig:class_distribution}
\end{figure}

\subsection{Phase III: Post-Client Pruning}
In this phase, we proceed with further refinement of the global model after removing the identified noisy clients in Phase II. This refinement entails conducting normal federated training for an additional $T_{post}$ rounds. The fundamental distinction here lies in the composition of clients involved in the federated training process. Specifically, only the remaining un-pruned clients participate in this phase.

To elaborate, we adopt the standard federated learning approach, such as vanilla FedAvg or any other FL optimizer of choice, to fine-tune the global model over the un-pruned clients' datasets. This process ensures that the global model continues to benefit from the diverse data contributions of clean clients while excluding the potentially disruptive influence of noisy clients. Algorithm~\ref{alg:clipfl} presents the details of our proposed method.

%% file: secs/experiments.tex
\section{Experiments} \label{sec:experiments}
In this section, we conduct a series of experiments to empirically evaluate the effectiveness of our proposed method. Our objective is to address the following research questions:

\begin{itemize}
\item \textbf{RQ1:} How does ClipFL compare to SOTA FL optimizers without pruning?
\item \textbf{RQ2:} How does ClipFL compare to SOTA FL methods designed to handle noisy labels?
\item \textbf{RQ3:} What are the impacts of ClipFL's hyper-parameters on its performance?
\item \textbf{RQ4:} How does the performance of ClipFL vary with different noise levels?
\bluetext{\item \textbf{RQ5:} How does the performance of ClipFL vary using synthetic validation dataset?}
\end{itemize}

We present comprehensive experimental results addressing each research question, shedding light on the efficacy and robustness of our proposed method. Specifically, we address RQ1 in Section~\ref{RQ1}, RQ2 in Section~\ref{RQ2}, RQ3 in Section~\ref{RQ3}, RQ4 in Section~\ref{RQ4}, \bluetext{and RQ5 in Section~\ref{RQ5}.}

\subsection{Experiment Settings} \label{exp_settings}

\noindent\textbf{FL Setting.}
We simulate a federated setting of 100 clients, where 50 of them are randomly selected to have their labels flipped. For the pre-client pruning phase (phase I), we set communication rounds to 80, i.e. $T_{pre}=80$, and number of clients to be considered as clean to 5, i.e $m=5$, unless specified otherwise. For the client pruning phase (phase II), we set the client pruning percentage to $50\%$, i.e. $p = 0.5$, unless specified otherwise. Lastly, for the post-client pruning phase (phase III), we set the communication rounds to 40, i.e. $T_{post}=40$. Therefore, the total communication rounds is $120$. The client sample rate is fixed to $0.1$, i.e $C=0.1$ during the entire federation process and phases.

For local training, we set the number of local epochs to $10$, and local batch size to $10$. We use an SGD optimizer with a learning rate of $0.03$ and $0.9$ momentum with no weight decay. Following~\cite{dosovitskiy2021image}, we use cross-entropy with label smoothing loss with softmax temperature $10.0$ and label smoothing coefficient $0.1$. We resize the input images to $224\times224$ and do not use any data augmentation techniques. \\

\noindent\textbf{Dataset and Model.}
We use CIFAR-10 and CIFAR-100~\cite{cifar} datasets for our evaluation. Both CIFAR-10 and CIFAR-100 are fully labeled popular computer vision datasets containing 60,000 samples each. Each data sample is a $32\times32$ RGB image. CIFAR-10 contains 10 different objects, and CIFAR-100 contains 100 objects. We split each dataset into train, validation and test with 9:1:2 ratio respectively.\footnote{Train, validation and test split sizes are 45,000, 5000, and 10,000 respectively.} We partition the train split across all clients, give the validation split to the central server for server-side client model evaluation, and also leave the test split on the server for global model performance evaluation. Following recent practices in using transformers in FL~\cite{qu2022rethinking}, we use the Vision Transformer~\cite{dosovitskiy2021image} architecture for its successful adoptions in computer vision tasks. The ViT-Tiny~\cite{steiner2022train} variant from the TIMM library\footnote{\url{https://github.com/huggingface/pytorch-image-models}} with a patch size of 16 (5.5M parameters) was used for our experiments for its small size that reduces communication costs of sending and receiving model weights and computation costs on edge devices. \\

\newcolumntype{g}{>{\columncolor{gray!25}}c}
\begin{table} 
\centering
\caption{Comparison of performance results between vanilla FL optimizers (without pruning) and those augmented with ClipFL (with pruning).}
\label{tab:main}
\resizebox{1.0\linewidth}{!}{
\begin{tabular}{cg|gggg|gggg} \toprule 
\rowcolor{white} & Partitioning & \multicolumn{4}{c|}{IID} & \multicolumn{4}{c}{NIID ($\alpha=0.5$)} \\
\rowcolor{white} Baseline & Dataset                            & \multicolumn{2}{c}{CIFAR10} & \multicolumn{2}{c|}{CIFAR100} & \multicolumn{2}{c}{CIFAR10} & \multicolumn{2}{c}{CIFAR100} \\ 
\rowcolor{white} & Noise Rate                                                        & $    \mu=0.5 $ & $\mu=0.8$ & $\mu=0.5$ & $\mu=0.8$ & $\mu=0.5$ & $\mu=0.8$ & $\mu=0.5$ & $\mu=0.8$ \\ \midrule
\rowcolor{white} \multirow{3}{*}{FedAvg~\cite{mcmahan2017communication}} & Vanilla   & $    55.64\% $ & $42.18\%$      & $42.06\%$      & $31.71\%$      & $43.37\%$      & $34.59\%$      & $40.16\%$      & $31.35\%$ \\ 
                                                                         & +ClipFL   &$\bm{69.42\%}$  & $\bm{59.14\%}$ & $\bm{43.20\%}$ & $\bm{41.21}\%$ & $\bm{49.56\%}$ & $\bm{48.34\%}$ & $\bm{42.74\%}$ & $\bm{37.02\%}$ \\ 
\rowcolor{white}                                                         & $\Delta$  &$+13.78\%$      &$+16.96\%$      &$+ 1.14\%$      &$+ 9.50\%$      &$+ 6.19\%$      &$+13.75\%$      &$ +2.25\%$      &$+ 5.67\%$ \\ \midrule
\rowcolor{white} \multirow{3}{*}{FedProx~\cite{li2020federated}}         & Vanilla   & $58.64\%$      & $45.11\%$      & $\bm{41.16\%}$ & $29.42\%$      & $46.25\%$      & $33.50\%$      & $38.48\%$      & $29.37\%$ \\ 
                                                                         & +ClipFL   & $\bm{68.99\%}$ & $\bm{68.92\%}$ & $39.86\%$      & $\bm{45.13\%}$ & $\bm{54.85\%}$ & $\bm{53.93\%}$ & $\bm{40.81\%}$ & $\bm{40.16\%}$ \\ 
\rowcolor{white}                                                         & $\Delta$  &$+10.35\%$      &$+23.81\%$      &$- 1.30\%$      &$+15.71\%$      &$+ 8.60\%$      &$+20.43\%$      &$+ 2.33\%$      &$+10.79\%$ \\ \midrule
\rowcolor{white} \multirow{3}{*}{FedNova~\cite{wang2020tackling}}        & Vanilla   & $56.55\%$ & $40.62\%$ & $43.67\%$ & $32.11\%$ & $46.86\%$ & $34.25\%$ & $39.85\%$ & $29.56\%$ \\ 
                                                                         & +ClipFL   & $\bm{67.48\%}$ & $\bm{62.00\%}$ & $\bm{47.23\%}$ & $\bm{39.71\%}$ & $\bm{53.03\%}$ & $\bm{49.82\%}$ & $\bm{43.69\%}$ & $\bm{34.72\%}$ \\ 
\rowcolor{white}                                                         & $\Delta$  &$+10.93\%$ &$+21.38\%$ &$+ 3.56\%$ &$+ 7.60\%$ &$+ 6.17\%$ &$+15.57\%$ &$+ 3.84\%$ &$+ 5.16\%$ \\ \midrule
\rowcolor{white} \multirow{3}{*}{FedAdam~\cite{reddi2021adaptive}}       & Vanilla   & $53.97\%$ & $42.35\%$ & $\bm{38.08\%}$ & $22.51\%$ & $\bm{41.39\%}$ & $30.88\%$ & $32.33\%$ & $23.50\%$ \\ 
                                                                         & +ClipFL   & $\bm{68.99\%}$ & $\bm{61.16\%}$ & $35.35\%$ & $\bm{34.14\%}$ & $40.39\%$ & $\bm{49.84\%}$ & $\bm{36.27\%}$ & $\bm{34.36\%}$ \\ 
\rowcolor{white}                                                         & $\Delta$  &$+15.02\%$ &$+18.81\%$ &$- 2.73\%$ &$+11.63\%$ &$- 1.00\%$ &$+18.96\%$ &$+ 3.94\%$ &$+10.86\%$ \\ \bottomrule

\end{tabular}
}
\end{table}

\noindent\textbf{Data partitioning and Label Noise.}
We consider two data partitions for our evaluation: IID and Non-IID. For IID data partitioning, we uniformly distribute the dataset into $100$ same-size partitions and assign one to each client. For Non-IID data partitioning, we follow prior studies and use the Dirichlet distribution method \cite{hsu2019measuring}. Specifically, for each class $k$, we randomly draw $z_k \sim Dir_N(\alpha)$ from the Dirichlet distribution with a concentration parameter $\alpha$. We then assign $z_{k,j}$ portion of class $k$ to client $j$. In our experiment, we use $\alpha=0.5$ for Non-IID data partitioning. 

After partitioning the dataset across all clients, we randomly select \bluetext{$\rho$ ratio} of the clients to make their dataset labels noisy. \bluetext{We set $\rho=0.5$ in our experiments for RQ1-3. This choice reflects a balanced
scenario where half of the clients own noisy data, providing a robust evaluation of ClipFL’s performance in response to RQ1-3} \bluetext{For noise model}, we use symmetric label noise modeling to make the selected clients data mislabeled, where every sample's label is randomly flipped to a different label from before according to a symmetric noise transition matrix with \bluetext{noise level $\mu$}. \bluetext{This choice is firmly grounded in the literature on machine learning with noisy labels, where the symmetric noise model is widely adopted due to its challenging nature and its ability to provide a consistent benchmark for evaluation without introducing bias toward specific implementations of noise matrices.} We evaluate our method under two different noise levels: low noise level ($\mu=0.5$) and high noise level ($\mu=0.8$). Figure~\ref{fig:class_distribution} visualizes an example of the class distribution of clients with and without label noise for IID and Non-IID data partitioning. \\

\noindent\textbf{Evaluation Metric.} We evaluate the global model on the test dataset and report the average performance over the last $10$ communication rounds in our experimentation, following~\cite{morafah2023practical, collins2021exploiting}. \\

\noindent\textbf{Implementation}
We implement our experiment code using FedZoo-Bench\footnote{\url{https://github.com/MMorafah/FedZoo-Bench}}~\cite{morafah2023practical} in PyTorch~\cite{paszke2019pytorch}. The ViT-Tiny~\cite{steiner2022train} model is adopted from the Timm library~\cite{rw2019timm}. The implementations of the baselines FedAvg~\cite{mcmahan2017communication}, FedProx~\cite{li2020federated}, FedNova~\cite{wang2020tackling}, and FedAdam~\cite{reddi2021adaptive} were provided by FedZoo-Bench, with little modifications. The details of implementations of FedCorr~\cite{xu2022fedcorr}, FedLSR~\cite{jiang2022towards}, and RHFL~\cite{fang2022rhfl} are provided in section \ref{sec:implementation}.  We make our code publicly available at~\url{https://github.com/MMorafah/ClipFL}.

\subsection{RQ1: Comparison with SOTA FL optimizers} \label{RQ1}
To address RQ1, we conduct a comprehensive comparative analysis between our proposed method and several SOTA FL optimizers. The objective is to assess the advantages of our pruning approach over SOTA FL optimizers, which typically perform federated learning over the entire client population, including both noisy and clean ones. Specifically, we consider four prominent FL optimizers: FedAvg \cite{mcmahan2017communication}, FedProx \cite{li2020federated}, FedNova \cite{wang2020tackling}, and FedAdam \cite{reddi2020adaptive}. Following recommendations by~\citet{nguyen2022begin} 
and~\citet{chen2023on}, to initialize models with pre-trained weights that enhance the FL optimizers' performance, we use weights pre-trained on ImageNet \cite{5206848} as the initial weights for all experiments in this section. \\

\begin{table}
\centering
\caption{Noisy Client Identification Accuracy of ClipFL with different FL optimizers.}
\label{tab:pruneacc}
\resizebox{1.0\linewidth}{!}{
\begin{tabular}{cg|gggg|gggg} \toprule 
\rowcolor{white} & Partitioning & \multicolumn{4}{c|}{IID} & \multicolumn{4}{c}{Non-IID} \\
\rowcolor{white} Baseline & Dataset & \multicolumn{2}{c}{CIFAR10} & \multicolumn{2}{c|}{CIFAR100} & \multicolumn{2}{c}{CIFAR10} & \multicolumn{2}{c}{CIFAR100} \\ 
\rowcolor{white} & Noise Level & $\mu=0.5$ & $\mu=0.8$ & $\mu=0.5$ & $\mu=0.8$ & $\mu=0.5$ & $\mu=0.8$ & $\mu=0.5$ & $\mu=0.8$ \\ \midrule
\rowcolor{white} \multicolumn{2}{c|}{FedAvg~\cite{mcmahan2017communication} + ClipFL}  & $98\%$ & $94\%$ & $94\%$ & $92\%$ & $72\%$ & $80\%$ & $84\%$ & $90\%$ \\ \midrule
\rowcolor{white} \multicolumn{2}{c|}{FedProx~\cite{li2020federated} + ClipFL} & $94\%$ & $94\%$ & $98\%$ & $98\%$ & $74\%$ & $86\%$ & $86\%$ & $92\%$ \\ \midrule
\rowcolor{white} \multicolumn{2}{c|}{FedNova~\cite{wang2020tackling} + ClipFL} & $96\%$ & $96\%$ & $96\%$ & $94\%$ & $66\%$ & $82\%$ & $92\%$ & $92\%$ \\ \midrule
\rowcolor{white} \multicolumn{2}{c|}{FedAdam~\cite{reddi2021adaptive} + ClipFL} & $94\%$ & $96\%$ & $88\%$ & $92\%$ & $74\%$ & $92\%$ & $94\%$ & $92\%$ \\ \bottomrule
\end{tabular}
}
\end{table}

\noindent\textbf{Performance Analysis.} Table~\ref{tab:main} presents the performance comparison between vanilla SOTA FL optimizers without pruning and those augmented with ClipFL. Notably, we observe a significant enhancement in performance across various scenarios when integrating ClipFL with different FL optimizers. For instance, under IID data partitioning and high noise level ($\mu=0.8$), the performance improves by at least $16\%$ for CIFAR-10 and at least $7\%$ for CIFAR-100. Across Non-IID and IID data partitioning, the performance improvement of ClipFL is similarly pronounced. Additionally, the performance improvement is more substantial under higher noise level ($\mu=0.8$), underscoring the efficacy of our approach.

When comparing the performance of ClipFL across different FL optimizers, FedProx generally outperformed others with ClipFL, achieving the highest test accuracy in 5 out of 8 cases. This highlights the compatibility and effectiveness of FedProx when combined with our proposed pruning approach under noisy labels scenario. \\

\noindent\textbf{Noisy Client Identification Analysis.} Table~\ref{tab:pruneacc} presents the accuracy of identifiying noisy clients of ClipFL, defined as the ratio of correctly identified noisy clients to the total clients marked as noisy.\footnote{$\text{Noisy Client Identification Accuracy} = \frac{\text{correctly identified noisy clients}}{\text{total clients marked as noisy}}$.}

As it is evident, ClipFL demonstrates high accuracy in identifying noisy clients, achieving a minimum accuracy of 88\% under IID data partitioning. However, we observe a decrease in accuracy identifying noisy clients under Non-IID data partitioning compared to IID scenarios. This decline can be attributed to the increased difficulty in discerning whether a drop in validation accuracy is due to noisy data or data heterogeneity in the Non-IID case. Despite these challenges, ClipFL maintains a commendable accuracy, achieving at least 66\% accuracy in identifying noisy clients under the most challenging scenario of Non-IID data partitioning with a noise level of $\mu=0.5$ for the CIFAR-10 dataset. Furthermore, ClipFL achieves at least 80\% accuracy in other Non-IID scenarios.

Interestingly, despite the lower noisy client identification accuracy in the most challenging Non-IID scenario (CIFAR-10, $\mu=0.5$), ClipFL still generally outperforms vanilla FL optimizers. This observation underscores the effectiveness of our approach in protecting clean clients from being adversely affected by noisy data, ultimately leading to performance improvements. \\

\begin{figure}
    \centering
    \begin{minipage}[t]{0.48\textwidth}
        \vspace{0pt}
        \includegraphics[width=1.0\textwidth]{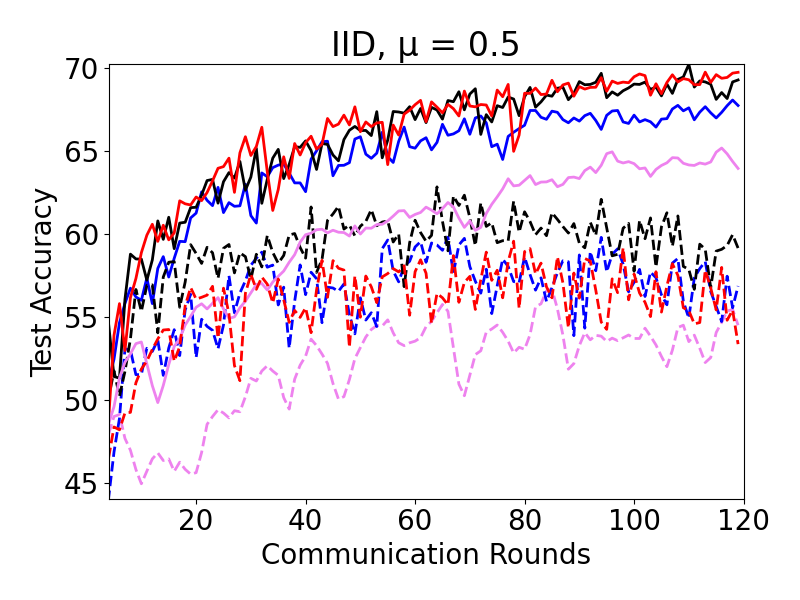}\\
    \end{minipage}%
    \begin{minipage}[t]{0.48\textwidth}
        \vspace{0pt}
        \includegraphics[width=1.0\textwidth]{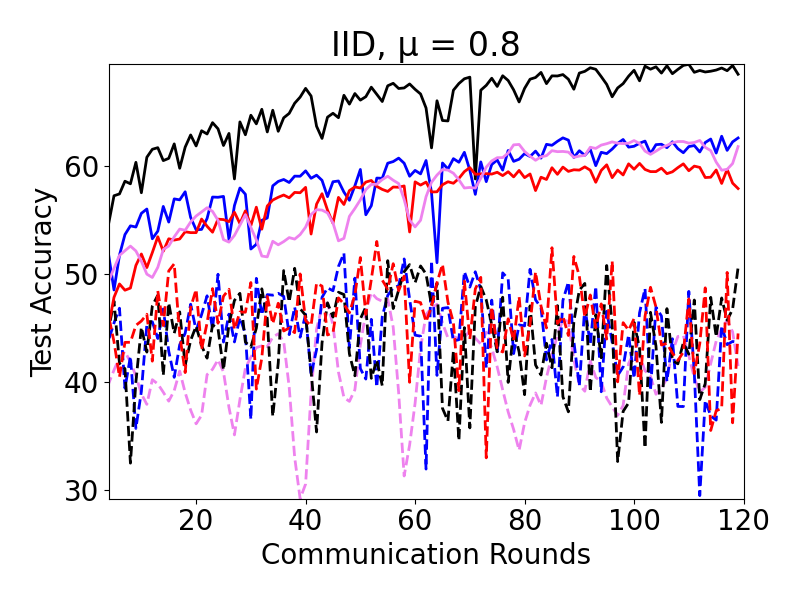}\\
    \end{minipage}%
    
    \begin{minipage}[t]{0.48\textwidth}
        \vspace{0pt}
        \includegraphics[width=1.0\textwidth]{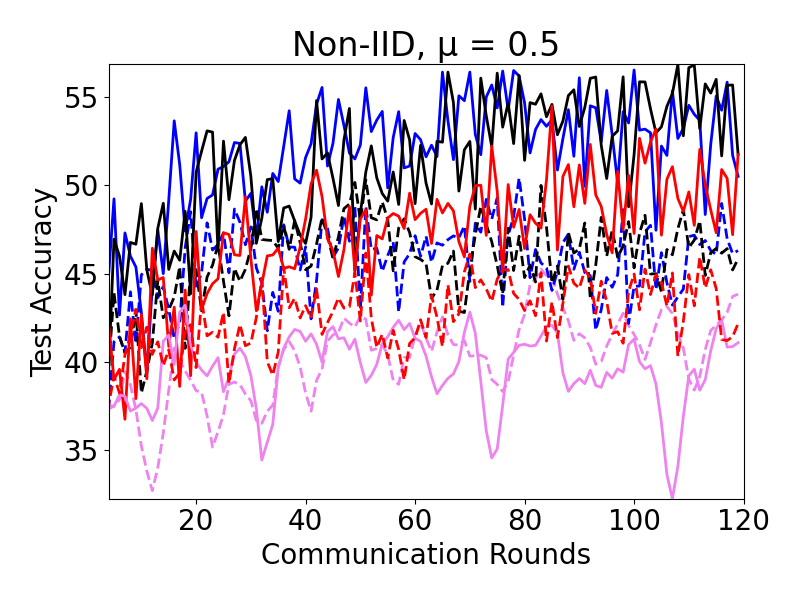}\\
    \end{minipage}%
    \begin{minipage}[t]{0.48\textwidth}
        \vspace{0pt}
        \includegraphics[width=1.0\textwidth]{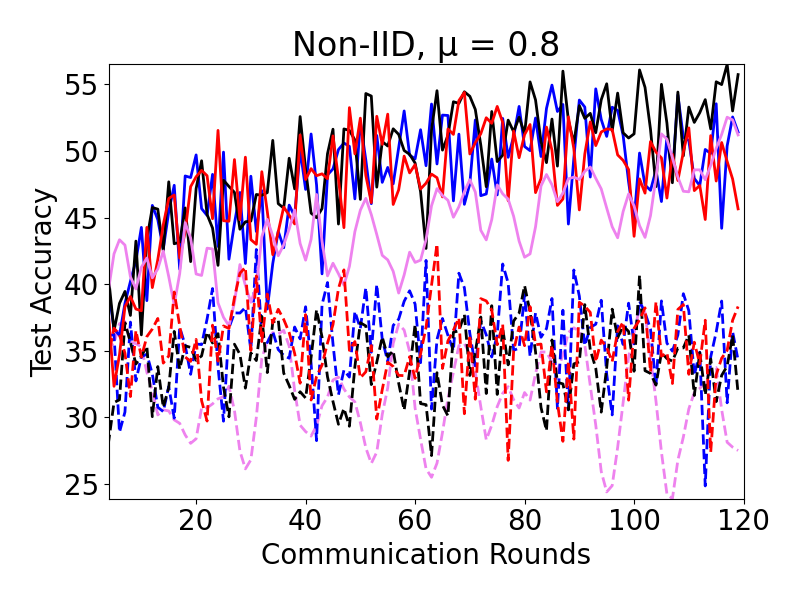}\\
    \end{minipage}%

    \begin{minipage}[t]{0.8\textwidth}
        \centering
        \includegraphics[width=1.0\textwidth]{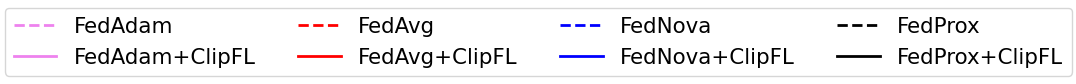}\\
    \end{minipage}%
    
    \caption{Test accuracy versus communication rounds for different vanilla FL optimizers and ClipFL on the CIFAR-10 dataset. The top row displays results for IID data partitioning, while the bottom row displays results for Non-IID data partitioning.}
    \label{fig:convergence_cifar10}
\end{figure}

\begin{figure}[h]
    \centering
    \begin{minipage}[t]{0.48\textwidth}
        \vspace{0pt}
        \includegraphics[width=1.0\textwidth]{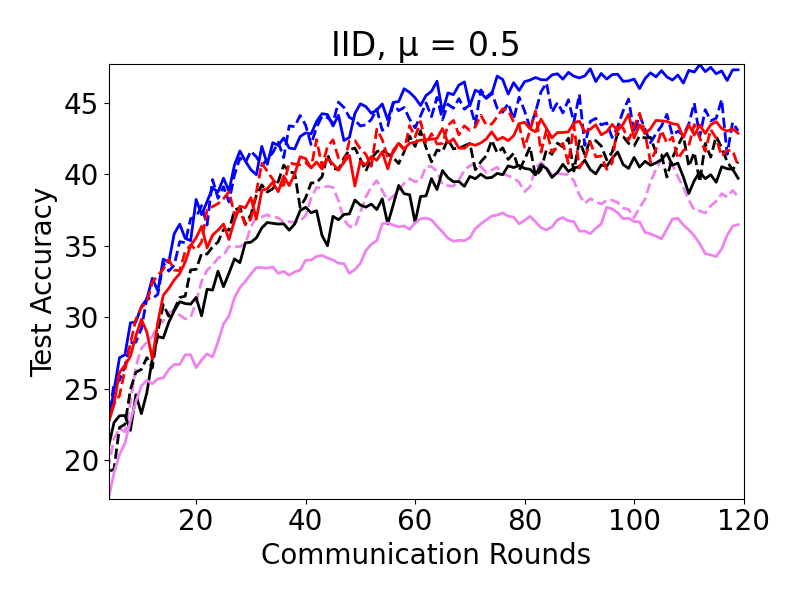}\\
    \end{minipage}%
    \begin{minipage}[t]{0.48\textwidth}
        \vspace{0pt}
        \includegraphics[width=1.0\textwidth]{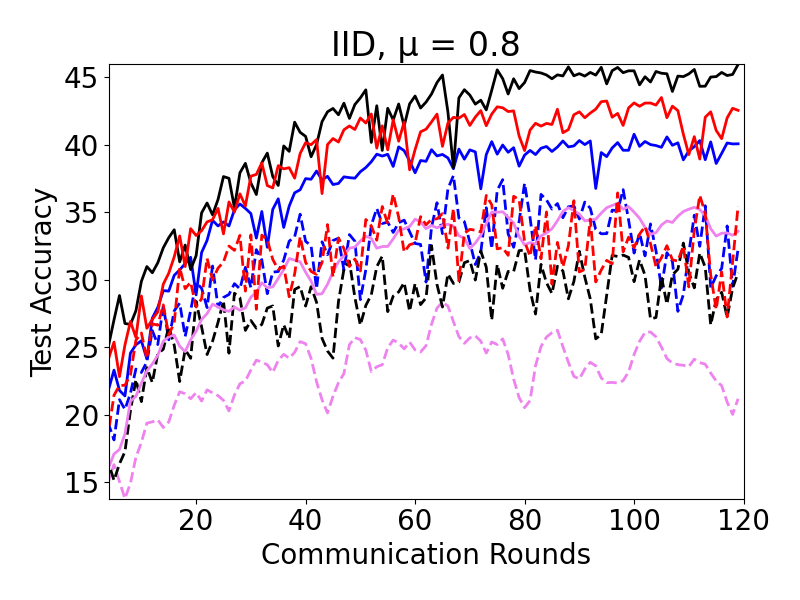}\\
    \end{minipage}%
    
    \begin{minipage}[t]{0.48\textwidth}
        \vspace{0pt}
        \includegraphics[width=1.0\textwidth]{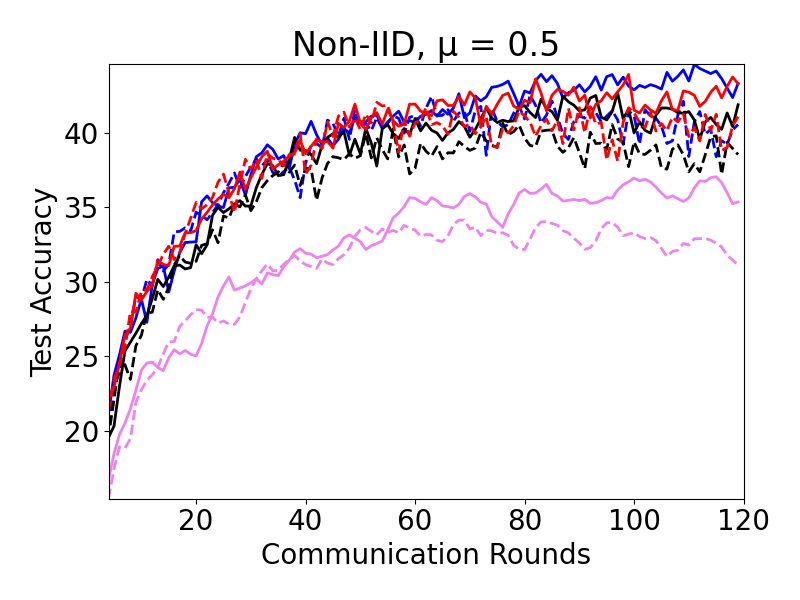}\\
    \end{minipage}%
    \begin{minipage}[t]{0.48\textwidth}
        \vspace{0pt}
        \includegraphics[width=1.0\textwidth]{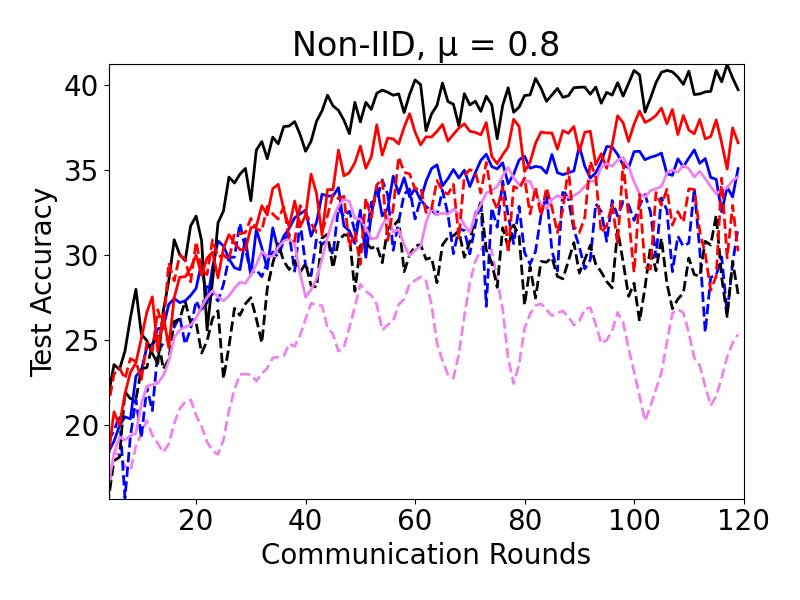}\\
    \end{minipage}%
    \hspace{0.15\textwidth}

    \begin{minipage}[t]{0.8\textwidth}
        \centering
        \includegraphics[width=1.0\textwidth]{images/Mar14_plots/legend.png}\\
    \end{minipage}%
    
    \caption{Test accuracy versus communication rounds for different vanilla FL optimizers and ClipFL on the CIFAR-100 dataset. The top row displays results for IID data partitioning, while the bottom row displays results for Non-IID data partitioning.}
    \label{fig:convergence_cifar100}
\end{figure}

\textbf{Convergence Analysis.} Figure~\ref{fig:convergence_cifar10} and Figure~\ref{fig:convergence_cifar100} illustrate the accuracy achieved in each communication round for the CIFAR-10 and CIFAR-100 datasets, respectively. It is evident that ClipFL consistently and significantly improves the performance of each FL optimizer baseline, with only a few instances where performance is comparable. Moreover, while the performance of vanilla FL optimizers shows minimal improvement over communication rounds, the performance of ClipFL consistently improves. Additionally, ClipFL reduces the overall communication cost of FL methods by identifying noisy clients and reducing the client pool size, resulting in a 17\% reduction in overall communication cost compared to each FL optimizer baseline over the total 120 communication rounds. These findings underscore the efficacy and efficiency of ClipFL in achieving improved convergence and reduced communication overhead compared to vanilla FL optimizers.

\subsection{RQ2: Comparison with SOTA FL methods that deal with noisy labels}\label{RQ2}
To address RQ2, we rigorously compare our proposed method with three state-of-the-art federated learning (FL) methods designed to handle noisy labels: FedCorr~\cite{xu2022fedcorr}, RHFL~\cite{fang2022rhfl}, and FedLSR~\cite{jiang2022towards}. \\

\noindent\textbf{FL with Noisy Labels Baselines and Settings.} FedCorr is a multi-stage FL framework that identifies noisy clients and corrects their noisy samples using global model predictions. RHFL addresses the noisy client challenge by employing symmetric cross-entropy loss for local training and re-weights clients based on their confidence. FedLSR focuses on regularizing local training to prevent over-fitting on noisy samples. We adopt the same experimental setup as detailed in Section \ref{RQ1}. To ensure fairness, we initialize models with random weights for all methods, adhering to each baseline's original implementation. FedAvg serves as the base FL optimizer for ClipFL in our experiments in this section. Further details on the hyperparameters used for the baselines are discussed in Section \ref{sec:hyperparameters}.

\begin{table}
\centering
\caption{Performance comparison between SOTA FL with noisy labels methods and ClipFL. $--$ marks diverged results in this table.}
\label{tab:comp_fl_noisy}
\resizebox{1.0\linewidth}{!}{
\begin{tabular}{g|gggg|gggg} \toprule 
\rowcolor{white} Partitioning &                                       \multicolumn{4}{c|}{IID}          &                   \multicolumn{4}{c}{Non-IID} \\
\rowcolor{white} Dataset & \multicolumn{2}{c}{CIFAR10} & \multicolumn{2}{c|}{CIFAR100} & \multicolumn{2}{c}{CIFAR10} & \multicolumn{2}{c}{CIFAR100} \\
\rowcolor{white} Noise level                           & $\mu=0.5$    & $\mu=0.8$    & $\mu=0.5$    & $\mu=0.8$    & $\mu=0.5$    & $\mu=0.8$     & $\mu=0.5$    & $\mu=0.8$     \\ \midrule
\rowcolor{white} FedAvg~\cite{mcmahan2017communication}& $42.16\%$    & $35.25\%$    &$\bm{21.06\%}$& $16.99\%$    & $34.48\%$    & $26.23$       & $17.92\%$    & $14.64\%$     \\ \midrule
\rowcolor{white} FedCorr~\cite{xu2022fedcorr}          & $47.10\%$    & $41.32\%$    & $20.22\%$    & $16.81\%$    & $35.79\%$    & $\bm{39.10\%}$& $18.33\%$    & $14.88\%$     \\
\rowcolor{white} RHFL~\cite{fang2022rhfl}              & $23.53\%$    & $19.82\%$    &  $4.83\%$    &  $3.96\%$    & $19.39\%$    & $18.07\%$     &  $4.64\%$    &  $4.04\%$     \\ 
\rowcolor{white} FedLSR~\cite{jiang2022towards}        & $--$         &$--$          &$--$          &$--$          &$--$          &$--$           &$--$          &$--$           \\
                 ClipFL                                &$\bm{49.48\%}$&$\bm{48.73\%}$&$20.66\%$     &$\bm{20.31\%}$&$\bm{38.62\%}$&$38.50\%$     &$\bm{19.12\%}$&$\bm{18.36\%}$ \\ \bottomrule
\end{tabular}
}
\end{table}

\begin{figure}
    \centering
    \begin{minipage}[t]{0.48\textwidth}
        \vspace{0pt}
        \includegraphics[width=1.0\textwidth]{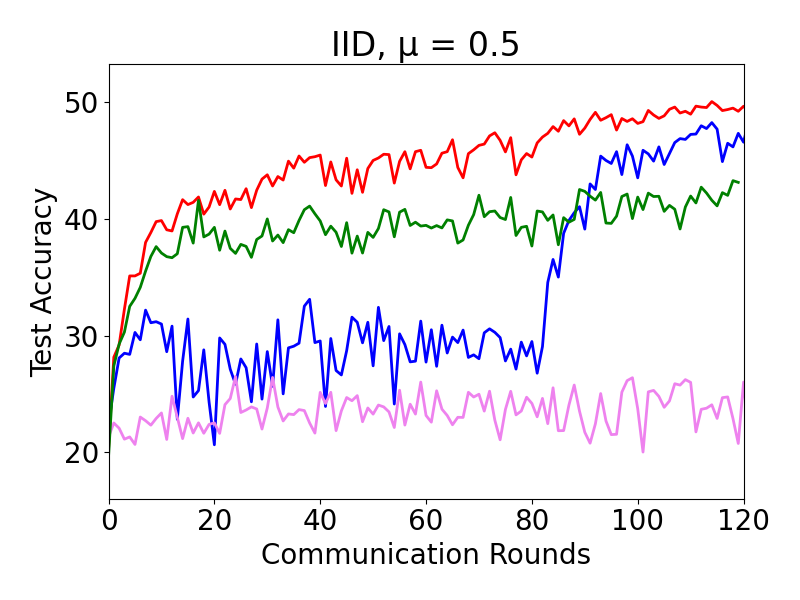}\\
    \end{minipage}%
    \begin{minipage}[t]{0.48\textwidth}
        \vspace{0pt}
        \includegraphics[width=1.0\textwidth]{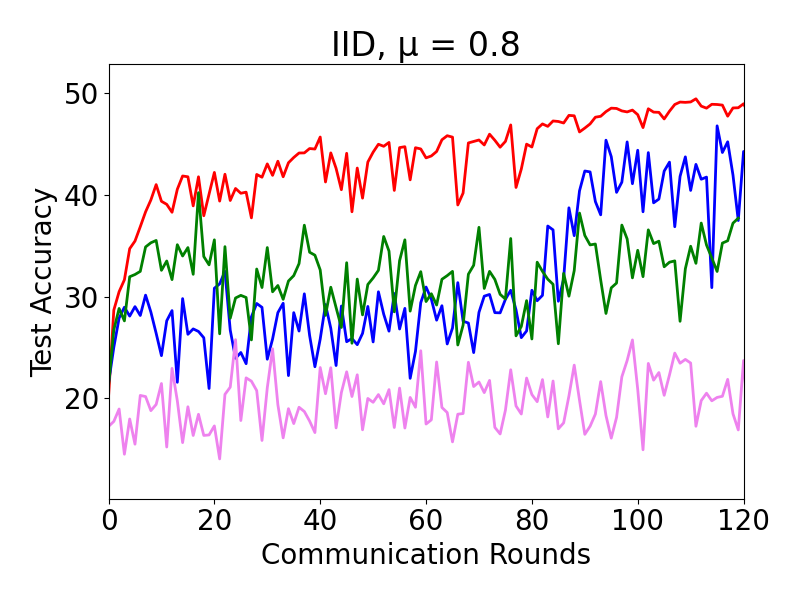}\\
    \end{minipage}%
    
    \begin{minipage}[t]{0.48\textwidth}
        \vspace{0pt}
        \includegraphics[width=1.0\textwidth]{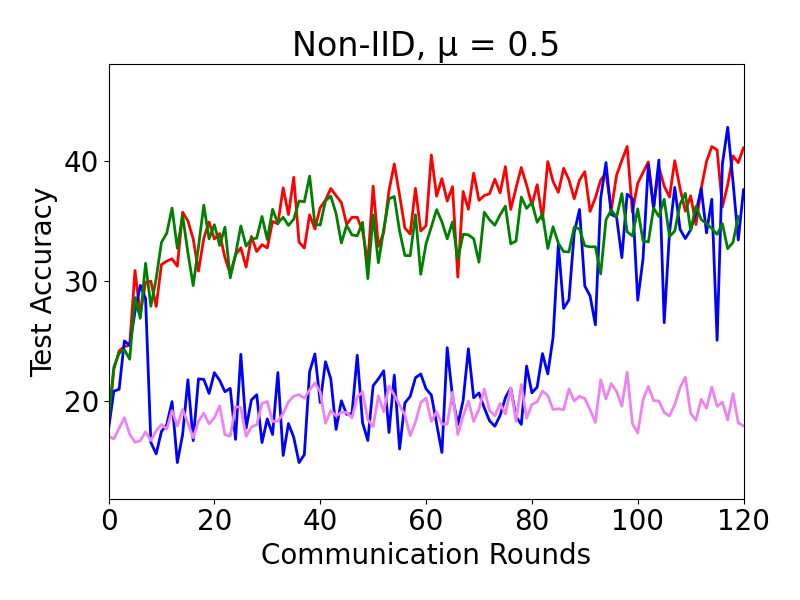}\\
    \end{minipage}%
    \begin{minipage}[t]{0.48\textwidth}
        \vspace{0pt}
        \includegraphics[width=1.0\textwidth]{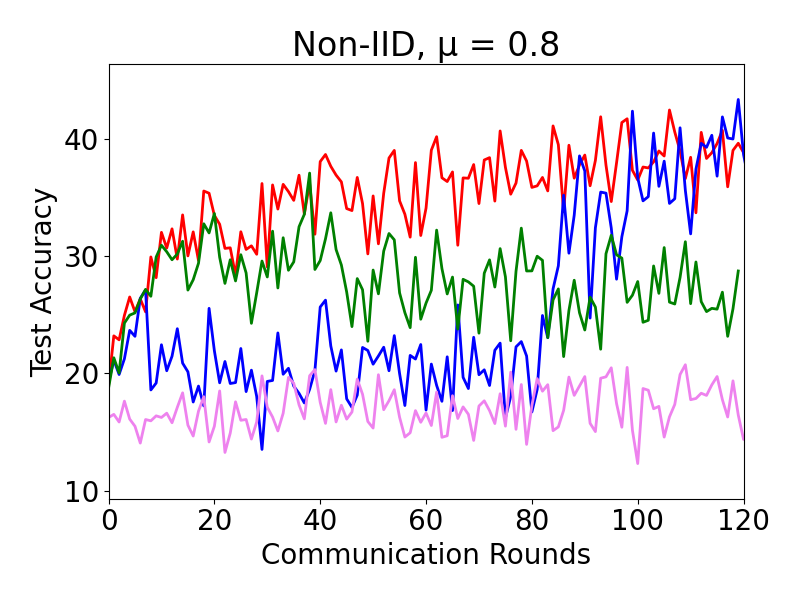}\\
    \end{minipage}%

    \begin{minipage}[t]{0.6\textwidth}
        \centering
        \includegraphics[width=1.0\textwidth]{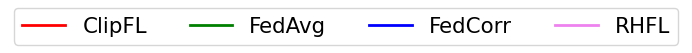}\\
    \end{minipage}%
    
    \caption{Test accuracy versus communication rounds for different vanilla FL optimizers and ClipFL on the CIFAR-10 dataset. The top row displays results for IID data partitioning, while the bottom row displays results for Non-IID data partitioning.}
    \label{fig:convergence_fl_noisy_c10}
\end{figure}

\begin{figure}
    \centering
    \begin{minipage}[t]{0.48\textwidth}
        \vspace{0pt}
        \includegraphics[width=1.0\textwidth]{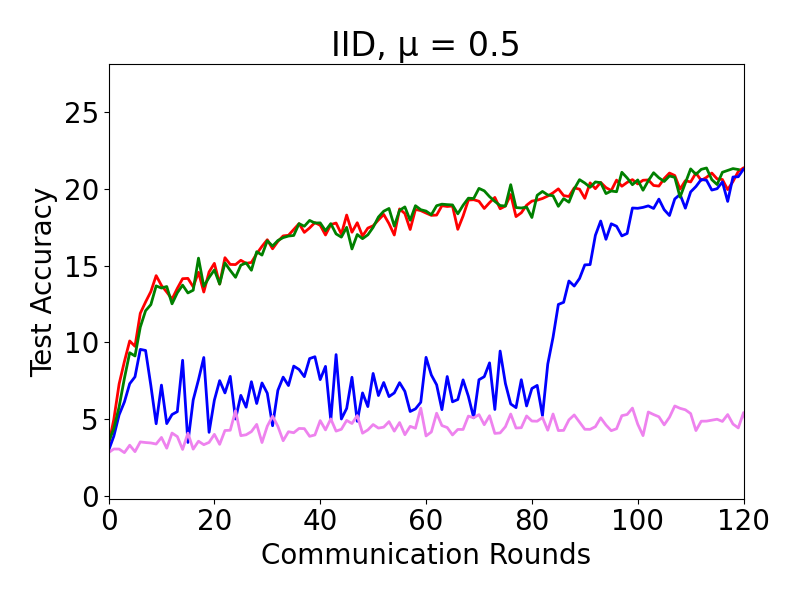}\\
    \end{minipage}%
    \begin{minipage}[t]{0.48\textwidth}
        \vspace{0pt}
        \includegraphics[width=1.0\textwidth]{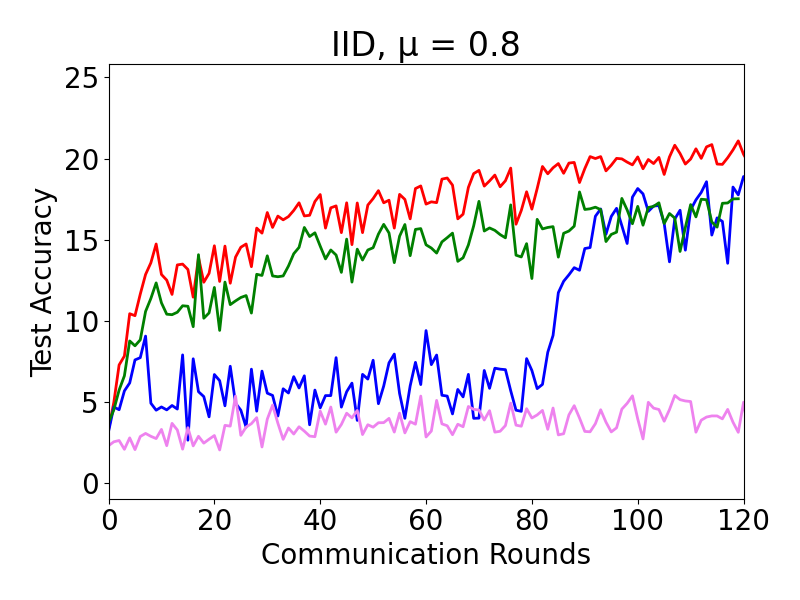}\\
    \end{minipage}%
    
    \begin{minipage}[t]{0.48\textwidth}
        \vspace{0pt}
        \includegraphics[width=1.0\textwidth]{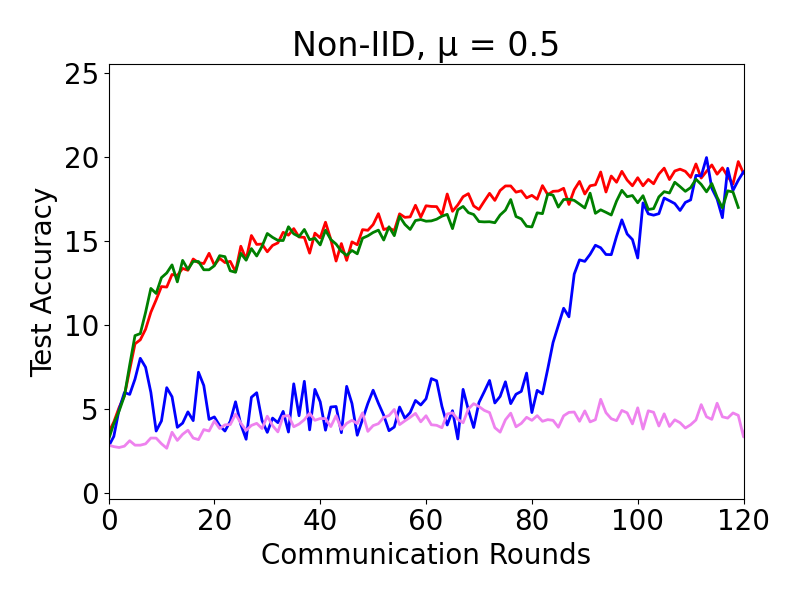}\\
    \end{minipage}%
    \begin{minipage}[t]{0.48\textwidth}
        \vspace{0pt}
        \includegraphics[width=1.0\textwidth]{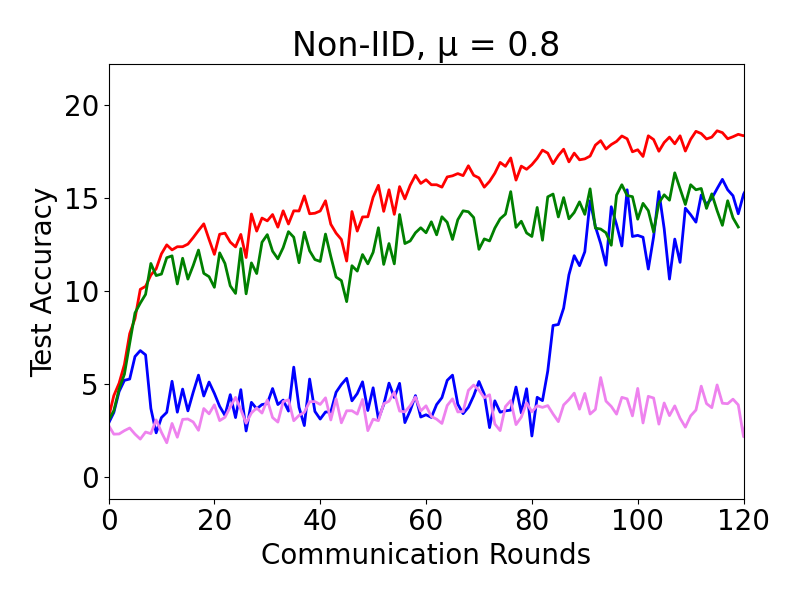}\\
    \end{minipage}%

    \begin{minipage}[t]{0.5\textwidth}
        \centering
        \includegraphics[width=1.0\textwidth]{images/sota_plots/legend.png}\\
    \end{minipage}%
    
    \caption{Test accuracy versus communication rounds for different vanilla FL optimizers and ClipFL on the CIFAR-100 dataset. The top row displays results for IID data partitioning, while the bottom row displays results for Non-IID data partitioning.}
    \label{fig:convergence_fl_noisy_c100}
\end{figure}

\noindent \textbf{Performance Analysis.} Table~\ref{tab:comp_fl_noisy} presents the performance comparison between the baselines and ClipFL. Overall, ClipFL consistently outperforms the baselines and SOTA performance, except for CIFAR-10 (IID, $\mu=0.5$) and CIFAR-100 (Non-IID, $\mu=0.8$) cases. Notably, on CIFAR-10 with IID and \bluetext{noise level $\mu=0.8$}, ClipFL exhibits a remarkable improvement of 7\% over the best-performing baseline, i.e. FedCorr. Similarly, on CIFAR-100 with Non-IID and \bluetext{noise level $\mu=0.8$}, ClipFL surpasses the SOTA performance by 4\%. Several key observations emerge: (1) Interestingly, while FedLSR and RHFL employ strategies to address noisy labels through robust local training and client re-weighting, they fail to achieve competitive performance, especially under higher noise levels. FedLSR fails to converge, noted as $--$ entries in the table. Moreover, RHFL exhibits poor performance, particularly on the CIFAR-100 dataset. (2) In contrast to RHFL and FedLSR, FedCorr demonstrates relatively better performance. FedCorr focuses on correcting noisy labels, and our results validate the effectiveness of this approach. However, FedCorr's performance heavily relies on having a well-trained global model as a pseudo-labeler, which is challenging to obtain under higher noise levels. This dependency is evident in the notable decrease in FedCorr's performance for higher noise levels ($\mu=0.8$) compared to lower noise levels ($\mu=0.5$), highlighting the challenge of relying solely on accurate global models for label correction. For example, on CIFAR-10 IID, FedCorr's performance drops by 6\% under high noise level ($\mu=0.8$) compared to low noise level ($\mu=0.5$). (3) Surprisingly, vanilla FedAvg demonstrates better performance compared to two prominent FL methods for noisy labels, RHFL and FedLSR, across all cases. Moreover, FedAvg achieves the best performance for CIFAR-100 (IID, $\mu=0.5$) and its performance closely rivals that of FedCorr in most cases under Non-IID data partitioning settings, even outperforming FedCorr for CIFAR-100 (IID, $\mu=0.8$). These findings highlight the performance issues in SOTA FL methods for noisy labels and underscore the need for designing robust methods capable of operating effectively in noisy label environments within FL frameworks.

\bluetext{It is noteworthy to highlight that while removing noisy data might seem to lead to information loss, it is crucial to recognize that
this loss pertains to misleading and noisy information rather than useful and informative signal. ClipFL’s approach of filtering out noisy clients and data ensures that the global model is trained on more reliable data, leading to improved performance. While an ideal approach would involve recovering the true labels from noisy data, this process typically requires a well-trained and reliable global model—a challenging feat in federated learning, especially under realistic
conditions involving data heterogeneity and a high prevalence of noisy data. ClipFL’s approach of filtering out noisy clients and data ensures that the global model is trained on more reliable and useful data, leading to improved performance. By leveraging a small public dataset, ClipFL circumvents the difficulties associated with achieving a reliable global model, thereby providing a more effective solution to the label correction problem in federated learning.  The validation dataset provides a consistent reference point, which allows for more effective handling
of clients with noisy labels compared to relying solely on a well-trained global model.}
These findings underscore the superiority of ClipFL in effectively addressing label noise without the need for a well-performing global model as a pseudo-labeler, leading to improved performance across various noisy label scenarios. \\

\noindent\textbf{Convergence Analysis} Figures \ref{fig:convergence_fl_noisy_c10} and \ref{fig:convergence_fl_noisy_c100} present the accuracy versus communication round plots for CIFAR-10 and CIFAR-100 datasets, respectively. Across different cases, ClipFL consistently outperforms the baselines as communication rounds progress, except for CIFAR-100 (IID, $\mu=0.5$). Notably, at higher noise levels ($\mu=0.8$), ClipFL exhibits more substantial performance improvements compared to the baselines.

Several insights can be made from these figures: (1) We observe that RHFL's performance remains relatively static across communication rounds, indicating limited adaptability to noisy label scenarios. (2) Conversely, FedCorr initially exhibits poor performance until approximately round 80 when label correction occurs. This underscores the necessity for warm-up rounds to establish a well-trained global model for effective label correction in FedCorr. Following label correction at round 80, FedCorr shows rapid performance improvement, particularly notable in the $\mu=0.5$ scenario, where it approaches ClipFL's performance. However, at high noise levels ($\mu=0.8$), FedCorr's performance gains post-correction are slower. (3) Interestingly, vanilla FedAvg demonstrates consistent improvement over communication rounds and exhibits rapid convergence in the majority of cases, unlike FL with noisy labels baselines. Furthermore, FedAvg performs very closely to ClipFL on CIFAR-100 (IID, $\mu=0.5$) and eventually surpasses ClipFL during the final rounds. These observations further support our findings and analysis in Table~\ref{tab:comp_fl_noisy}, highlighting the efficacy issues in SOTA FL methods for noisy labels and the necessity for designing robust FL methods in noisy label environments.

In contrast to previous approaches, \textbf{\textit{ClipFL demonstrates a distinct advantage in fast convergence, making it suitable for scenarios with limited communication rounds.}} Additionally, ClipFL reduces overall communication costs by 17\% over 120 rounds compared to the baseline methods. This efficiency underscores the practical viability of ClipFL in federated learning setups with noisy labels.


\begin{figure}[t]
    \centering
    \begin{minipage}{0.35\textwidth}
    \centering
        \includegraphics[width=\textwidth]{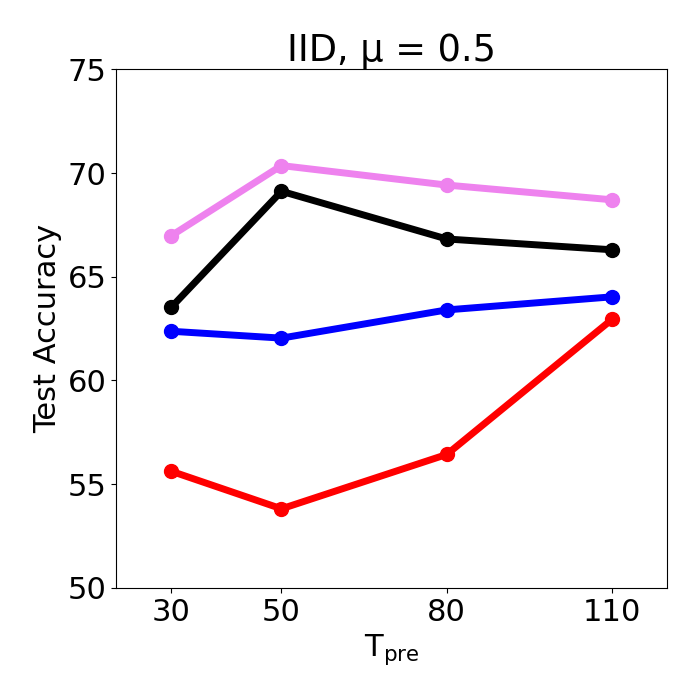}\\
    \end{minipage}%
    \hspace{0.01\textwidth}
    \begin{minipage}{0.35\textwidth}
    \centering
        \includegraphics[width=\textwidth]{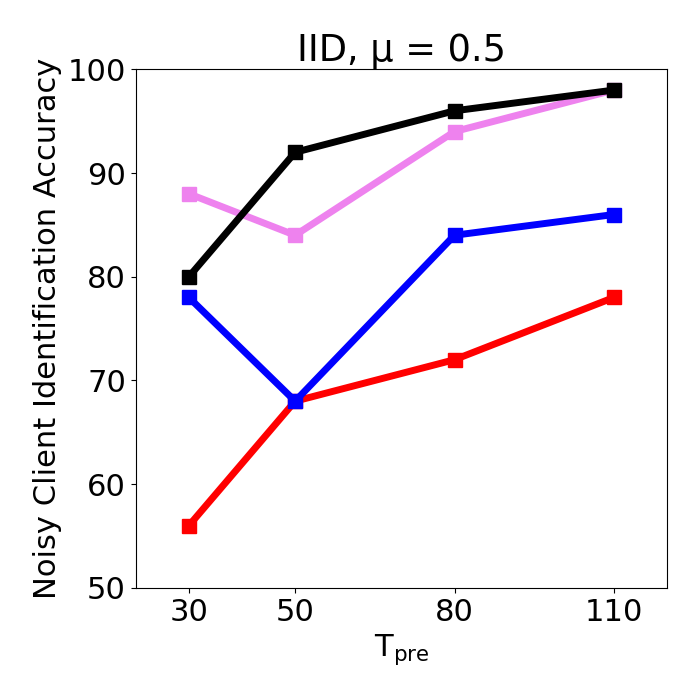}\\
    \end{minipage}%
    \hspace{0.01\textwidth}
    \begin{minipage}{0.1\textwidth}
    \centering
        \includegraphics[width=\textwidth]{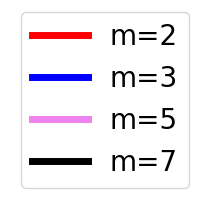}\\
    \end{minipage}%

    \begin{minipage}{0.35\textwidth}
    \centering
        \includegraphics[width=\textwidth]{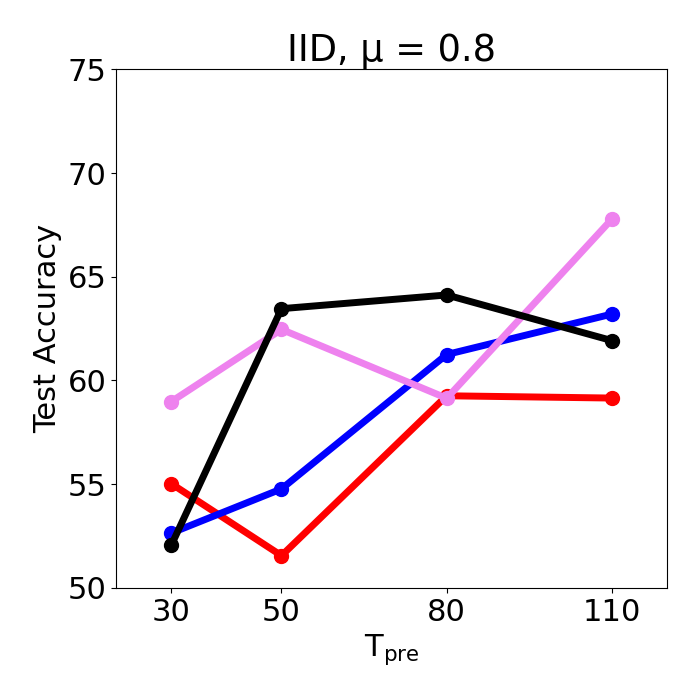}\\
    \end{minipage}%
    \hspace{0.01\textwidth}
    \begin{minipage}{0.35\textwidth}
    \centering
        \includegraphics[width=\textwidth]{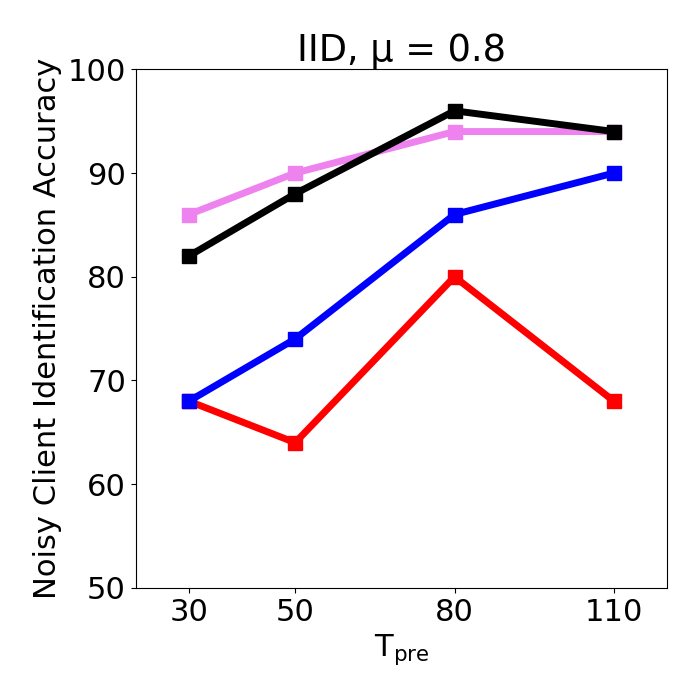}\\
    \end{minipage}%
    \hspace{0.11\textwidth}
    \caption{Effect of pre-client pruning communication rounds ($T_{pre}$) and number of clients considered as clean ($m$) on the performance of ClipFL. The left figures show the test accuracy and the right figures show the corresponding noisy client identification accuracy on CIFAR-10 IID data partitioning on CIFAR-10 (IID, $\mu=0.5$).}
    \label{fig:abla_hp_plots}
\end{figure}

\subsection{RQ3: Impact of ClipFL's hyper-parameters} \label{RQ3}

In this section, we thoroughly investigate the impact of ClipFL's hyper-parameters on its performance. Specifically, we analyze the effects of varying three key hyper-parameters: (1) Pre-client pruning communication rounds ($T_{pre}$), (2) the number of clients considered as clean based on their validation accuracy ($m$), and (3) the percentage of clients pruned ($p$). We conduct our experiments on CIFAR-10 IID data partitioning in this section and the setting is the same as described in Section~\ref{exp_settings}. It is also noteworthy to mention that 50\% of the clients are noisy in this setting. \\

\noindent\textbf{Effect of $T_{pre}$ and $m$.} Figure~\ref{fig:abla_hp_plots} illustrates the influence of different pre-client pruning communication rounds $T_{pre}$ and $m$ on test accuracy performance and noisy client identification accuracy for low and high noise levels, i.e. $\mu=0.5$ and $\mu=0.8$, respectively. Regarding $T_{pre}$, we observe that smaller values negatively impact the accuracy of noisy client identification. This occurs because ClipFL has insufficient historical data for each client, leading to less robust noisy client identification. Consequently, lower test accuracy is observed, particularly evident with $T_{pre}=30$, which exhibits the lowest performance and noisy client identification accuracy. Conversely, higher values of $T_{pre}$ result in better performance and more accurate noisy client identification due to increased historical client data.

Examining different values of $m$, we find that $m=2$ yields the poorest performance across all scenarios. This suggests that considering only the top 2 clients as clean while labeling the rest as noisy is overly aggressive. However, with $m=2$, performance improves as pre-client pruning communication rounds ($T_{pre}$) increase. Optimal trade-offs between performance and pre-client pruning communication rounds are observed with $m=3$ and $m=5$, which provide a balanced approach between aggressiveness and performance. Specifically, $m=3$ and $m=5$ exhibit satisfactory performance with $T_{pre}=50$ and $T_{pre}=80$. \\

\begin{figure}[t]
    \centering
    \begin{minipage}{0.4\textwidth}
    \centering
        \includegraphics[width=\textwidth]{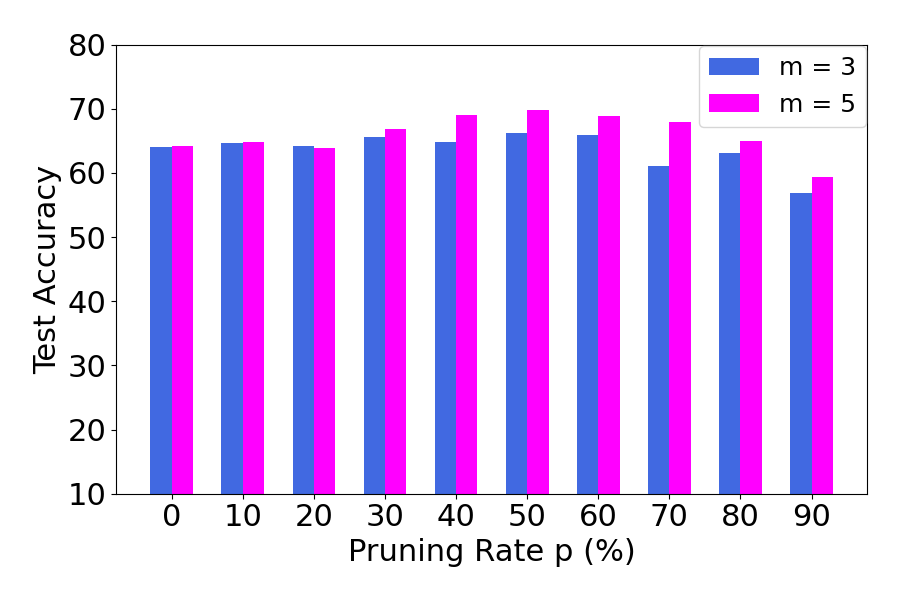}\\
    \end{minipage}%
    \caption{Effect of pruning rate ($p$) on the performance of ClipFL on CIFAR-10 (IID, $\mu=0.5$).}
    \label{fig:abla_pr_plots}
\end{figure}

\noindent\textbf{Effect of $p$.}  Figure~\ref{fig:abla_pr_plots} illustrates the impact of different pruning rates on performance for $m=3$ and $m=5$, with $T_{pre}=80$ fixed, on CIFAR-10 (IID, $\mu=0.5$). Observing the results, we find that there is no significant difference in performance between $m=3$ and $m=5$ for pruning rates below 30\%. The most notable difference occurs around optimal pruning percentages, specifically at $p=40\%$, $50\%$, and $60\%$. However, as the pruning percentage surpasses 60\%, a general decline in performance is observed for both $m=3$ and $m=5$.

Comparing the bars for $m=3$ and $m=5$, we consistently observe that the $m=5$ bar is higher than the $m=3$ bar, except under low pruning percentages. This discrepancy arises because $m=5$ offers more accurate identification of noisy clients, leading to superior performance improvements. This assertion is supported by the top plots in Figure~\ref{fig:abla_hp_plots}, where $m=5$ consistently outperforms $m=3$ in terms of more accurate identification of noisy clients.

\subsection{RQ4: How does the performance of ClipFL vary with different noise levels?} \label{RQ4}
In this section, we comprehensively investigate the performance of ClipFL under varying noise levels to address RQ4. Our experimentation focuses on the CIFAR-10 dataset with IID partitioning.

\begin{figure}[h]
    \centering
    \begin{minipage}{0.4\textwidth}
    \centering
        \includegraphics[width=\textwidth]{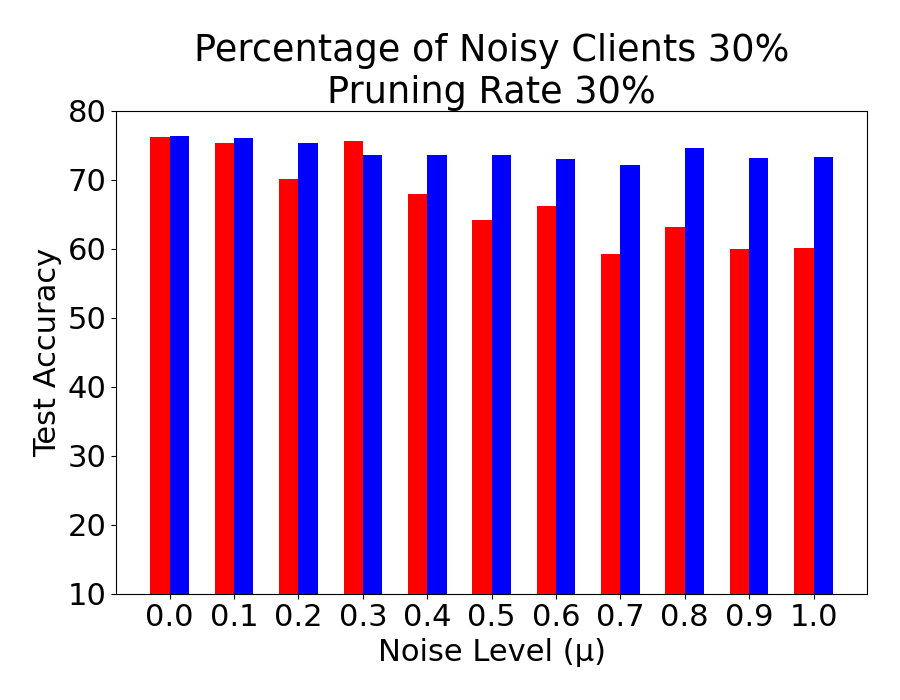}\\
    \end{minipage}%
    \hspace{0.01\textwidth}
    \begin{minipage}{0.4\textwidth}
    \centering
        \includegraphics[width=\textwidth]{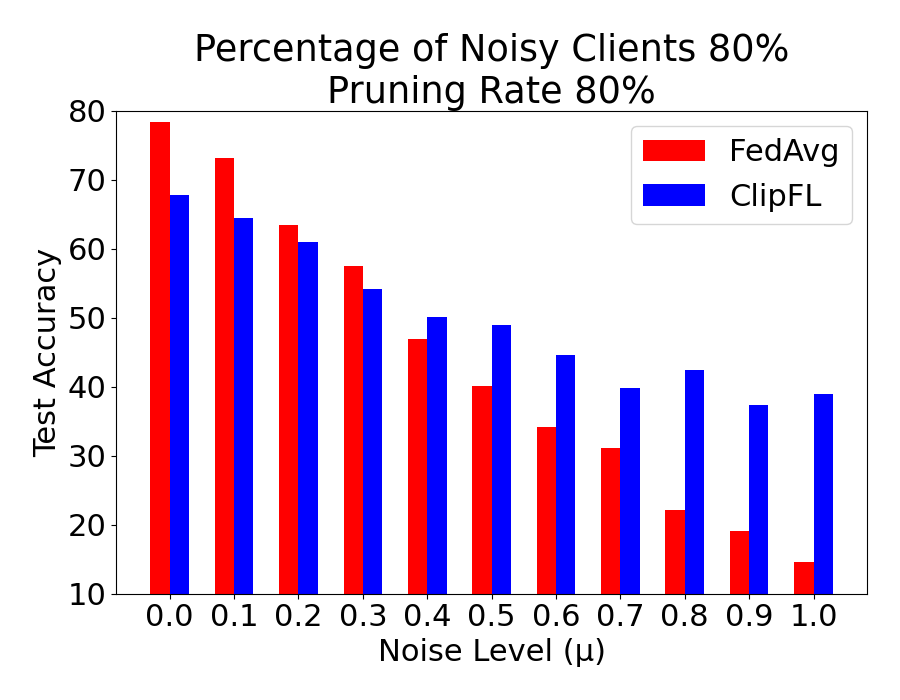}\\
    \end{minipage}%
    \caption{Performance of ClipFL under different \bluetext{noise levels} on CIFAR-10 dataset with IID partitioning.}
    \label{fig:abla_ns_plots}
\end{figure}

Figure~\ref{fig:abla_ns_plots} illustrates the performance of vanilla FedAvg and FedAvg+ClipFL across different noise levels ($\mu$) for two different percentages of noisy clients. We employ a pruning rate of $p=30\%$ for the 30\% noisy client percentage and $p=80\%$ for the 80\% noisy client percentage.

Overall, we observe that as noise levels elevate, ClipFL's performance enhancement becomes increasingly pronounced. Specifically, ClipFL exhibits greater performance for noise levels $\mu \geq 0.4$, whereas vanilla FedAvg demonstrates superior performance for low noise levels ($\mu < 0.3$). This observation suggests that in scenarios with minimal noise (e.g., $\mu=0.1, 0.2$), the presence of noisy clients does not significantly impair performance; thus pruning is less advantageous. Particularly for the 80\% noisy client scenario, vanilla FedAvg consistently outperforms ClipFL for $\mu \leq 0.3$.

Conversely, as noise levels escalate to $\mu \geq 0.8$, ClipFL substantially enhances performance, demonstrating at least an 11\% improvement for the 30\% noisy client scenario and at least 18\% enhancement for the 80\% noisy client scenario. This substantial performance improvement underscores the effectiveness of ClipFL, especially in environments fraught with substantial \bluetext{noise levels}.

These findings underscore the critical role of pruning in mitigating the detrimental impact of noisy clients, particularly in settings characterized by significant \bluetext{noise levels}. While pruning may yield marginal or no benefits under very low noise levels, its efficacy becomes increasingly pronounced as noise levels surge. This observation spotlights the utility of ClipFL in scenarios where label noise prevails, facilitating substantial performance gains.

\bluetext{
\subsection{RQ5: How does the performance of ClipFL vary using synthetic validation dataset?}\label{RQ5} 
In this section, we investigate the impact of the validation dataset on ClipFL's performance. In practice, different entities in various fields, such as the NIH and WHO in the healthcare domain, maintain the quality and annotation of publicly available datasets. Furthermore, the server has ample resources and can utilize platforms like Amazon Mechanical Turk to collect and annotate data, which can then be verified by domain experts to ensure high quality—a method successfully employed in creating datasets for various machine learning tasks. Moreover, recent advancements in generative AI have made it possible to synthesize high-quality images using Diffusion models, which can be used for various purposes~\cite{yang2023diffusion, cao2024survey, zhang2023text}. To further investigate the impact of having no access to a real, clean validation dataset on ClipFL's performance, we conducted an experiment using a synthetic dataset generated with an off-the-shelf state-of-the-art (SOTA) text-to-image pre-trained Stable Diffusion model~\cite{rombach2022high}. Specifically, we utilized the ``CompVis/stable-diffusion-v1-4'' pre-trained Stable Diffusion model checkpoint from HuggingFace to generate the images, using 40 inference steps and a guidance scale of 7. In total, we synthesized 10,000 validation images, with an equal number of images for each class. We conducted our experiments using the same settings discussed in Section~\ref{RQ1}. \\} 

\bluetext{\noindent\textbf{Performance Analysis.} Table~\ref{tab:syn} presents the results using the synthetic dataset. We observe that, for the IID case, the maximum $\Delta$ difference between using real and synthetic data as the validation dataset is $-0.58\%$, while for the Non-IID Dir(0.5) case, it is $-0.65\%$. These results further highlight the practicality of ClipFL, even in the extreme scenario where there is no real validation dataset available and thus synthesized datasets are used. \\}

\bluetext{\noindent\textbf{Privacy Analysis.} It is important to note that our method does not necessitate the sharing of clients private data. The validation dataset is maintained solely at the server level and is used exclusively for model evaluation and hyper-parameter tuning, consistent with other common studies in FL. This approach aligns with privacy-preserving principles in FL, ensuring that clients' local data remains confidential throughout the process. Furthermore, synthesizing data at the server using Foundation models has been also investigated in FL without violating privacy of clients' raw data~\cite{abacha2024synthetic, morafah2024stable}. \\
}

\bluetext{\noindent\textbf{Comment on Size of Validation dataset.} It is crucial to emphasize that the size of the required validation dataset is typically much smaller than the combined size of all client datasets in FL, which makes the task of creating and maintaining a high-quality validation set more manageable. In our experiments, we observed that a validation set comprising just 1-5\% of the total federated dataset size was sufficient to achieve significant improvements in model performance.
}

\newcolumntype{g}{>{\columncolor{gray!25}}c}
\newcolumntype{s}{>{\columncolor{gray!25}}l}
\begin{table} 
\centering
\caption{\bluetext{Impact of using synthetic validation dataset generated using SOTA Stable Diffusion model on ClipFL's performance.}}
\label{tab:syn}
\resizebox{1.0\linewidth}{!}{
\begin{tabular}{s|gggg|gggg} \toprule 
\rowcolor{white} & \multicolumn{4}{c|}{IID} & \multicolumn{4}{c}{NIID ($\alpha=0.5$)} \\
\rowcolor{white} Baseline & \multicolumn{2}{c}{CIFAR10} & \multicolumn{2}{c|}{CIFAR100} & \multicolumn{2}{c}{CIFAR10} & \multicolumn{2}{c}{CIFAR100} \\ 
\rowcolor{white} & $    \mu=0.5 $ & $\mu=0.8$ & $\mu=0.5$ & $\mu=0.8$ & $\mu=0.5$ & $\mu=0.8$ & $\mu=0.5$ & $\mu=0.8$ \\ \midrule
\rowcolor{white} FedAvg (Vanilla)   & $    55.64\% $ & $42.18\%$      & $42.06\%$      & $31.71\%$      & $43.37\%$      & $34.59\%$      & $40.16\%$      & $31.35\%$ \\ 
\rowcolor{white} FedAvg+ClipFL (Real Data)   &$\bm{69.42\%}$  & $\bm{59.14\%}$ & $\bm{43.20\%}$ & $\bm{41.21}\%$ & $\bm{49.56\%}$ & $\bm{48.34\%}$ & $\bm{42.74\%}$ & $\bm{37.02\%}$ \\ 
\rowcolor{white} FedAvg+ClipFL (Synthetic Data)   &${69.16\%}$  & ${58.80\%}$ & ${42.83\%}$ & ${40.63}\%$ & ${49.24\%}$ & ${47.93\%}$ & ${42.26\%}$ & ${36.37\%}$ \\ 
\rowcolor{white} $\Delta$ (Real - Synthetic)  &$-0.26\%$      &$-0.34\%$      &$- 0.37\%$      &$-0.58\%$      &$-0.32\%$      &$-0.41\%$      &$-0.48\%$      &$-0.65\%$ \\  \bottomrule

\end{tabular}
}
\end{table}

\subsection{Summary and Discussion}
In our experimental study, we make the following key observations:
\begin{itemize}
    \item ClipFL significantly enhances the performance of SOTA FL optimizers under noisy label settings by effectively pruning noisy clients from the federation. This approach leads to more robust and accurate global model updates, contributing to higher overall accuracy.
    \item Traditional FL methods for noisy labels exhibit poor convergence and limited performance due to challenges in handling noisy clients. In contrast, ClipFL demonstrates fast convergence and substantial performance improvements over communication rounds. Its adaptability to noisy label scenarios ensures consistent progress in accuracy, even under challenging conditions.
    \item ClipFL exhibits robustness even in highly challenging scenarios with high noise levels. Its ability to identify and prune noisy clients effectively allows it to mitigate the adverse effects of noisy labels, ensuring stable and reliable performance across various noise levels.
    \item One of the notable advantages of ClipFL is its fast convergence and reduced communication overhead. By efficiently identifying and pruning noisy clients and reducing the number of communication rounds needed for convergence, ClipFL minimizes unnecessary communication and computational costs, making it an efficient solution for federated learning setups with noisy labels.
    \bluetext{\item ClipFL can be easily scaled to large-scale and severely noisy FL systems without the need
    for complex aggregation schemes (or advanced and complex FL methods) or extensive communication
    rounds to achieve a reliable global model.}
    \bluetext{\item While removing noisy data might seem to lead to information loss, it is crucial to recognize that this loss pertains to misleading and noisy information rather than useful and informative signal. ClipFL’s approach of filtering out noisy clients and data ensures that the global model is trained on more reliable and informative data, circumventing the challenges of achieving a well-trained and reliable global model in order to recover the true labels from noisy data, leading to improved performance. }
\end{itemize}

%% file: secs/hyperparameters.tex
\section{Implementation and Hyper-parameters} \label{sec:imp_hp}
In this section we discuss the details of our implementation and the hyper-parameters we used for each baseline in our experimentation.

\subsection{Implementations} \label{sec:implementation}
\noindent\textbf{FL optimizers implementation.} We adopt the implementations of FedAvg~\cite{mcmahan2017communication}, FedProx~\cite{li2020federated}, FedNova~\cite{wang2020tackling} and FedAdam~\cite{reddi2021adaptive} from FedZoo-Bench~\cite{morafah2023practical} with minimal adjustments.  

\noindent\textbf{FL with noisy labels methods implementation.} We implement FedCorr~\cite{xu2022fedcorr} using its original implementation\footnote{\url{https://github.com/Xu-Jingyi/FedCorr}} with no adjustments. We implement RHFL~\cite{fang2022rhfl} using their original implementation.\footnote{\url{https://github.com/FangXiuwen/Robust_FL}} We implement FedLSR~\cite{jiang2022towards} using their original implementation.\footnote{\url{https://github.com/Sprinter1999/FedLSR}} The experiments were conducted on a systme with two NVIDIA GeForce RTX 3090s.

\subsection{Hyper-parameters} \label{sec:hyperparameters}
In this part, we discuss the details of hyper-parameters we used in our experimentations. 

\noindent\textbf{Setting and default hyper-parameters.} 
We conducted extensive hyper-parameter tuning with different optimizers and learning rates using the ViT-Tiny~\cite{steiner2022train} model. We chose an SGD optimizer with a learning rate of 0.03 and momentum of 0.9.
Table \ref{tab:opt_hp} discusses the base hyper-parameters we used for all the baselines including FL optimizers and FL with noisy labels methods, unless specified otherwise. The pre-trained ViT-Tiny~\cite{steiner2022train} model has been downloaded from the TIMM library.\footnote{\url{https://github.com/huggingface/pytorch-image-models}} 
\begin{table}[h]
    \centering
    \caption{Loss and Optimizer Hyper-parameters} \label{tab:opt_hp}
    \begin{tabular}{c|c} 
    \toprule 
    Loss Type & Label Smoothing CE \\
    Label Smoothing Temperature & 10 \\
    Label Smoothing Level & 0.1 \\
    Optimizer Type & SGD \\
    Optimizer Learning Rate & 0.03 \\
    Optimizer Weight Decay & 0 \\
    Optimizer Momentum & 0.9 \\
    \bottomrule
    \end{tabular}
\end{table}

\noindent\textbf{FL optimizers' hyper-parameters.} For FedProx, we used $\mu=0.001$. For FedAdam we used $\tau=0.001$ and $0.01$ server learning rate. 

\noindent\textbf{FedCorr hyper-parameters.} We adjust the number of rounds per stage as 8 rounds for the pre-processing stage, 75 rounds for the fine-tuning stage, and 37 rounds for the normal federated learning stage. The relabel ratio was set to 0.5, confidence threshold to 0.5, $\alpha$ to 1, $\beta$ to 5, $k$ to 20, and clean set threshold set to 0.1. FedCorr's own loss uses MixUp~\cite{zhang2018mixup} and local proximal regularization, where both of which are techniques that boost the performance of FL algorithms in general in the presence of label noise. ClipFL does not use both techniques, so the Label Smoothing Cross Entropy loss is used instead. 

\noindent\textbf{RHFL hyper-parameters.} For the public dataset we use CIFAR-100 following their paper~\cite{fang2022rhfl}, with a size of 5000 samples. $\beta$ was set to 0.5. 

\noindent\textbf{FedLSR hyper-parameters.} FedLSR's warm-up rounds was set to 40 rounds and $\gamma$ set to 0.25, following \cite{jiang2022towards}'s results.

%% file: secs/conclusion.tex
\section{Conclusion} \label{sec:conclusion}
In this work, we introduced ClipFL, a novel approach for addressing the challenges of federated learning in the presence of noisy labels. By leveraging a systematic client pruning mechanism based on noise candidacy scores, ClipFL effectively identifies and removes noisy clients from the federation, leading to significant performance improvements and faster convergence compared to SOTA FL methods. Our experimental results demonstrate that ClipFL achieves robust and reliable performance even under highly challenging scenarios with varying noise levels and data heterogeneity.

Key advantages of ClipFL include its ability to enhance the performance of FL optimizers, mitigate the adverse effects of noisy labels, and reduce communication overhead. By prioritizing clean clients for model aggregation and pruning noisy clients effectively, ClipFL ensures stable and efficient federated learning across different datasets and noise conditions.

Overall, ClipFL presents a promising solution for federated learning applications where label noise is a prevalent challenge. Future research directions may include further exploration of ClipFL's performance under different noise distributions, investigation of adaptive client pruning strategies, and extension to other FL frameworks and applications.

\bluetext{Future work should focus on developing more advanced label correction methods that can accurately recover true labels from noisy data without degrading model performance.}
